\documentclass{sig-alternate}
\usepackage{mathptmx} 

\newcommand{\ignore}[1]{}
\usepackage{fancyhdr}
\usepackage[normalem]{ulem}
\usepackage[hyphens]{url}
\usepackage{microtype}
\usepackage{siunitx}
\usepackage{subfig}

\usepackage{listings}
\usepackage{amsmath}
\usepackage[normalem]{ulem}
\usepackage{graphicx}
\usepackage{xcolor}
\usepackage{soul}
\usepackage{threeparttable}

\usepackage[bookmarks=true,breaklinks=true,letterpaper=true,colorlinks,linkcolor=black,citecolor=blue,urlcolor=black]{hyperref}

\newcommand{\Sec}[1]{Section~\ref{#1}}
\newcommand{\Fig}[1]{Figure~\ref{#1}}
\newcommand{\Tbl}[1]{Table~\ref{#1}}

\renewcommand{\paragraph}[1]{\noindent\textbf{#1}}

\title{Benchmarking TPU, GPU, and CPU Platforms for Deep Learning}

\author{Yu (Emma) Wang, Gu-Yeon Wei and David Brooks\\
\textit{\{ywang03,gywei,dbrooks\}@g.harvard.edu}
\and
\textit{John A. Paulson School of Engineering and Applied Sciences}\\
\textit{Harvard University}
}

\begin{document}
\maketitle
\pagestyle{plain}

\begin{abstract}
Training deep learning models is compute-intensive and there is an industry-wide trend towards 
hardware specialization to improve performance. To systematically benchmark deep learning platforms, 
we introduce ParaDnn, a parameterized benchmark suite for deep learning that generates end-to-end models for fully connected (FC), convolutional (CNN), and recurrent (RNN) neural networks.
Along with six real-world models, we benchmark Google's Cloud TPU v2/v3, NVIDIA's V100 GPU, and an Intel Skylake CPU platform.
We take a deep dive into TPU architecture, reveal its bottlenecks, and highlight valuable lessons learned for future specialized system design.
We also provide a thorough comparison of the platforms and find that each
has unique strengths for some types of models. 
Finally, we quantify the rapid performance 
improvements that specialized software stacks provide for the TPU and GPU platforms. 


\end{abstract}

\section{Introduction}
Deep learning has revolutionized many application domains, defeating world champions in the game of Go~\cite{silver2017mastering},
surpassing humans in image classification~\cite{huang2017densely}, and
achieving competitive accuracy to humans in speech recognition~\cite{amodei2016deep} and language translation~\cite{wu2016google}, to name a few.
As such, there has been growing demand for new and better hardware and software platforms to support the training and deployment of even more sophisticated models. As researchers from both academia and industry scramble to propose and deploy new systems to meet this demand, there is a great need to concurrently develop a systematic and scientific approach to platform benchmarking. This benchmarking should not only compare performance of different platforms running a broad range of deep learning models, but also support deeper analysis of the interactions across the spectrum of different model attributes (e.g., hyperparameters), hardware design choices, and software support.   


Announced in May 2017, the Tensor Processing Unit (TPU) v2 is a custom ASIC.
Each TPU v2 device delivers a peak of \SI{180}{TFLOPS} on a single board.
TPU v3 was announced a year later and improves the peak performance to \SI{420}{TFLOPS}.
Cloud TPU became available for early academic access in February 2018. It is used in this paper.
The NVIDIA Tesla V100 Tensor Core is a Graphics Processing Unit (GPU) with the Volta architecture that was released in 2017.
CPUs have been found to be suitable for training in certain cases~\cite{hazelwood2018applied} and, therefore, are an important platform to include for comparison.
This study shows that no one platform is best for all scenarios. Different platforms offer advantages for different models based on their respective characteristics.
Moreover, given how rapidly deep learning models 
evolve and change, benchmarking must be updated continuously and run frequently.


Recent benchmarking efforts have been limited to relatively small collections of seemingly arbitrary DNN
models~\cite{mlperf,adolf2016fathom,chen2012benchnn,tao2018b}. Focusing on well-known models such 
as ResNet50~\cite{he2016deep} and Transformer~\cite{vaswani2017attention} can lead to misleading conclusions.
For example, Transformer is a large FC model that trains 3.5$\times$ faster on the TPU compared to the GPU; yet
focusing on this single model would not reveal the severe TPU memory bandwidth bottleneck that arises with FCs 
with more than 4k nodes. This highlights the risk of overly optimizing hardware and/or compilers 
for certain models. 


This paper proposes a collection of deep learning models (for training) created and curated to benchmark a set of state-of-the-art deep learning platforms.
In order to support broad and comprehensive benchmark studies, we introduce {\bf ParaDnn}, a parameterized deep learning 
benchmark suite. ParaDnn seamlessly generates thousands of parameterized multi-layer models, comprising 
fully-connected models (FC), convolutional neural networks (CNN), and recurrent neural networks (RNN).
ParaDnn allows systematic benchmarking across almost six orders-of-magnitude of model parameter size, exceeding
the range of existing benchmarks.

\begin{table*}[t]
\small
\begin{center}
\begin{threeparttable}
\begin{tabular}{|l|c|c|l|}

\multicolumn{1}{c}{Observation} & \multicolumn{1}{c}{ParaDnn\tnote{*}} & \multicolumn{1}{c}{Proof} &  \multicolumn{1}{c}{Insight/Explanation}\\

\hline

1. TPU does not exploit the parallelism from the model depth (layer count). & $\checkmark$ & Fig~\ref{fig:tpu_flops} & To design/upgrade new specialized systems, architects \\

\cline{1-3}


2. Many FC and CNN operations are bottlenecked by TPU memory bandwidth. & $\checkmark$ & Fig~\ref{fig:tpu_rooflines} & need to consider interactions between the operation  \\ 
\cline{1-3}


3. TPU suffers large overheads due to inter-chip communication bottlenecks. & $\checkmark$ & Fig~\ref{fig:tpu_multinode} & mix from key workloads (arithmetic intensity) and   \\ 
\cline{1-3}

4. TPU performance can be improved by $\ge$ 34\% by improving data infeed. & - & Fig~\ref{fig:casestudy} & system configurations (FLOPS, memory bandwidth/ \\ 

\cline{1-3}

5. TPU v3 optimizes compute-bound MatMuls by 2.3$\times$, memory-bound  & &  & capacity, and intra-chip and host-device interconnect). \\ 

~~~~ones by 3$\times$, and large embeddings by $> 3 \times$, compared to v2. & $\checkmark$ & Fig~\ref{fig:tpuv3} & TPU serves as a great example.\\

\hline
\hline

6. The largest FC models prefer CPU due to memory constraints. & $\checkmark$ & Fig~\ref{fig:fc_example} & Need for model parallelism on GPU and TPU.\\
\hline 

7. Models with large batch size prefer TPU. &  & Fig~\ref{fig:fc_speedup_tpu_gpu} & Large batches pack well on systolic arrays; \\
~~~~Those with small batch size prefer GPU. & - & Fig~\ref{fig:cnnrnn_speedup_tpu_gpu} & warp scheduling is flexible for small batches. \\
\hline 
8. Smaller FC models prefer TPU and larger FC models prefer GPU. & $\checkmark$ & Fig~\ref{fig:fc_speedup_tpu_gpu}  &  FC needs more memory bandwidth per core (GPU). \\
\hline 


9. TPU speedup over GPU increases with larger CNNs. & $\checkmark$ & Fig~\ref{fig:cnnrnn_speedup_tpu_gpu} &  TPU architecture is highly optimized for large CNNs. \\
\hline 

10. TPU achieves 2$\times$ (CNN) and 3$\times$ (RNN) FLOPS utilization compared to GPU.  & $\checkmark$ & Fig~\ref{fig:overall} & TPU is optimized for both CNN and RNN models. \\
\hline 


11. GPU performance scales better with RNN embedding size than TPU. & $\checkmark$ & Fig~\ref{fig:cnnrnn_speedup_tpu_gpu} & GPU is more flexible to parallelize non-MatMuls. \\
\hline
\hline

12. Within seven months, the software stack specialized for TPU  & & & It is easier to optimize for certain models\\
~~~~~~improved by up to 2.5$\times$ (CNN), 7$\times$ (FC), and 9.7$\times$ (RNN). & $\checkmark$ & Fig~\ref{fig:compiler} & than to benefit all models at once.\\
\hline


13. Quantization from 32 bits to 16 bits  &  &  Fig~\ref{fig:casestudy} & Smaller data types save memory traffic and enable \\
~~~~~~significantly improves TPU and GPU performance. & - & Fig~\ref{fig:compiler} & larger batch sizes, resulting in super-linear speedups. \\
\hline

14. TensorFlow and CUDA teams provide substantial performance &  &  & There is huge potential to optimize compilers 
\\
~~~~~~improvements in each update. & $\checkmark$ & Fig~\ref{fig:compiler} &  
even after the hardware has shipped.\\
\hline

\end{tabular}
\begin{tablenotes}
\item[*] Without ParaDnn the insights are not revealed, and/or lack deep explanations.
\end{tablenotes}
\end{threeparttable}
\end{center}
\vspace{-1em}
\caption{A summary of major observations and insights grouped by section of the paper.}
\label{table:insights}
\end{table*}

We combine these parameterized models with a collection of six real-world models, which serve as unique points within a broad spectrum of model attributes, 
to provide comprehensive benchmarking of hardware platforms. Table~\ref{table:insights}
summarizes fourteen observations and insights described throughout the paper that can inform future
domain-specific architecture, system, and software design.
We specifically mark the insights enabled by ParaDnn.
We start with a deep dive 
into the TPU v2 and v3 in Section~\ref{sec:tpu}, revealing architectural bottlenecks in computation capability, 
memory bandwidth, multi-chip overhead, and device-host balance (observations 1 through 5). Section~\ref{sec:xcompare} 
provides a comprehensive comparison of TPU and GPU performance, highlighting important differences between the two 
platforms (observations 6 through 11). The final three observations are detailed in Section~\ref{sec:compiler}, which explores
the performance improvements of specialized software stacks and quantized datatypes.


It is important to identify limitations of the study.
This paper highlights optimization opportunities in current architecture and system designs, as they provide valuable lessons for future design.
Optimization details are beyond its scope.
For example, the analysis focuses on training and not inference.
We do not study the performance of multi-GPU platforms or 256-node TPU systems, which may lead to different conclusions.
\Sec{sec:limitation} discusses these and other limitations of the study, which also motivate future work.


\section{Deep Learning Benchmarking}
\label{sec:synbench}

Recent success of deep learning (DL) has motivated development of benchmark suites, but existing suites have limitations.
There are two types, real-world benchmark suites such as MLPerf~\cite{mlperf},
Fathom~\cite{adolf2016fathom}, BenchNN~\cite{chen2012benchnn}, and BenchIP~\cite{tao2018b},
and micro-benchmark suites, such as DeepBench~\cite{deepbench} and BenchIP.
Each real-world suite contains a handful of popular DL models spanning a variety of model architectures.
Their limitation is that they only contain today's deep learning models, which may become obsolete as DL models evolve rapidly.
Further, they fail to reveal deep insights into interactions between DL model attributes and hardware performance, since
the benchmarks are sparse points in the vast space of deep learning models.
Micro-benchmark suites exercise basic operations (e.g., matrix multiplication or convolution) that are common in neural networks,
but they cannot simulate complex dependencies between different operations in end-to-end models.

To \textit{complement} existing benchmark suites for this study, we introduce ParaDnn, a parameterized benchmark suite for deep learning.\footnote{We plan to open-source ParaDnn.}
ParaDnn has the advantages of the above approaches,
with the goal of providing large 
``end-to-end'' models covering current and \textit{future} applications, and parameterizing the models to explore a much larger design space of DNN model attributes.
For example, a single end-to-end CNN model from ParaDnn contains 
a mixture of many different layers with different sizes of convolution, batch normalization, pooling, and FC layers. 
The complexity of ParaDnn workloads is comparable to that of real-world models (e.g., ResNet50 and Transformer), as will be shown 
in \Fig{fig:model_params}. Insights about hardware performance sensitivity to model attributes allow interpolating and extrapolating to future models of interest.
These insights could not be discovered with either the small point space 
exploration of the real-world benchmark suites or DeepBench's microbenchmarks,which do not capture inter-operation dependencies 
as ParaDnn does.




\subsection{ParaDnn Models}
ParaDnn includes end-to-end fully connected models (FC), convolutional neural networks (CNN), and recurrent neural networks (RNN).
The model types cover 95\% of Google's TPU workloads~\cite{jouppi2017datacenter},
all of Facebook's deep learning models~\cite{hazelwood2018applied},
and eight out of nine MLPerf models~\cite{mlperf} (with reinforcement (minigo) as an exception).
The image classification/detection and sentiment analysis models are CNNs;
the recommendation and translation models are FCs;
the RNN translator and another version of sentiment analysis are RNNs.
Speech recognition (DeepSpeech2) is a combination of CNN and GRU models.

\paragraph{Fully-Connected Models}
FC models comprise multiple fully-connected layers.
The architecture is
\begin{equation*}
\mbox{Input} \rightarrow [\mbox{Layer}[\mbox{Node}]] \rightarrow \mbox{Output},
\end{equation*}
where [Layer] means the number of layers is variable.
We can sweep the number of layers, the number of nodes per layer, and the numbers of input and output units of the datasets.

\paragraph{Convolutional Neural Networks}
CNN models are residual networks, the state-of-the-art model for image classification. 
The architecture of ParaDnn CNNs is
\begin{equation*}
\mbox{Input} \rightarrow [\mbox{Residual/Bottleneck Block}]\times 4 \rightarrow \mbox{FC} \rightarrow \mbox{Output}.
\end{equation*}
A residual network contains four groups of blocks~\cite{he2016deep}.
Each can be a residual block or a bottleneck block, followed by a fully-connected layer.
Residual blocks have two convolutional layers and two batch normalization layers, while bottleneck blocks have three of each.
Usually the minimum number of filters of a residual network is 64 and it doubles in every group, so the maximum is 512 filters.
We sweep the number of blocks per group, the minimum filters, and the datasets, 
including input images and number of categories as outputs.
An input image is square with three channels, represented by its length.
To keep the study tractable, 
we constrain each group to have the same number of blocks.

\paragraph{Recurrent Neural Networks}
RNNs comprise multiple layers of basic RNN, LSTM, or GRU cells as shown below.
\begin{equation*}
\mbox{Input} \rightarrow [\mbox{RNN/LSTM/GRU Cell}] \rightarrow \mbox{Output}.
\end{equation*}
Each token of the input sequence is embedded within a fixed length vector, and the length of the vector is the embedding size.
In ParaDnn, the number of layers and the embedding size are variable.
The variables in the dataset include the maximum length per input sequence and the vocabulary size.

\paragraph{Range of Hyperparameters and Datasets}
We choose the range of hyperparameters and datasets to cover the real models (\Sec{sec:exp:workload}), and we make sure the design space is reasonable.
Table~\ref{table:var_range} summarizes variables for each network type and how they are swept.
We also sweep training batch sizes.

\begin{table}[t]
   \centering
   \subfloat[Fully Connected Models\label{table:fc_var_range}]{
   \small
     \centering
     \begin{tabular}{|c|c|c|c|c|c|}

       \multicolumn{1}{c}{\textbf{Variable}}  & \multicolumn{1}{c}{Layer}  & \multicolumn{1}{c}{Nodes}  & \multicolumn{1}{c}{\emph{Input}} &\multicolumn{1}{c}{\emph{Output}}& \multicolumn{1}{c}{Batch Size} \\
\hline 
\textbf{Min} & 4 & 32  & 2000 & 200 & 64 \\
\hline 
\textbf{Max} & 128 & 8192 & 8000 & 1000 & 16384 \\
\hline 
\textbf{Inc} & $\times$2 & $\times$2 & +2000 & +200 & $\times$2 \\
\hline 
     \end{tabular}
     \vspace{-0.5em}
   }
   \vspace{-0.5em}
   \subfloat[Conv. Neural Nets: Residual and Bottleneck Blocks\label{table:cnn_var_range}]{
     \small
     \centering
      \begin{tabular}{|c|c|c|c|c|c|}
       \multicolumn{1}{c}{\textbf{Variable}}  & \multicolumn{1}{c}{Block}  &  \multicolumn{1}{c}{Filter}  &
 \multicolumn{1}{c}{\emph{Image}} & \multicolumn{1}{c}{\emph{Output}} &
  \multicolumn{1}{c}{Batch Size} \\
\hline 
\textbf{Min} & 1 & 16 &200 & 500 & 64 \\
\hline 
\textbf{Max} & 8 & 32 & 300 & 1500 & 1024 \\
\hline 
\textbf{Inc} & +1 & 64 & +50 & +500 & $\times$2 \\
\hline 
     \end{tabular}
     \vspace{-0.5em}
   }
   \vspace{-0.5em}
   \subfloat[Recurrent Neural Networks: RNN, LSTM, GRU\label{table:rnn_var_range}]{
     \small
     \centering
     \begin{tabular}{|c|c|c|c|c|c|}
       \multicolumn{1}{c}{\textbf{Variable}}  & \multicolumn{1}{c}{Layer}  &
  \multicolumn{1}{c}{Embed}   &
 \multicolumn{1}{c}{\emph{Length}} & \multicolumn{1}{c}{\emph{Vocab}} &
  \multicolumn{1}{c}{Batch Size} \\
\hline 
\textbf{Min} & 1 & 100 & 10 & 2 & 16\\
\hline 
\textbf{Max} & 13 & 900 & 90 & 1024 & 1024\\
\hline 
\textbf{Inc} & +4 & +400 & +40 & $\times$4 & $\times$4\\
\hline 
     \end{tabular}
   }
   \vspace{-0.5em}
   \caption{The ranges of the hyperparameters and dataset variables (\textit{italic}) chosen in this paper.}\label{table:var_range}
   \vspace{-1em}
\end{table}

\subsection{Real-World Models}
\label{sec:exp:workload}

In addition to ParaDnn, we include two of the three workloads written in TensorFlow from MLPerf~\cite{mlperf}, i.e.,
Transformer (translation)~\cite{vaswani2017attention} and ResNet-50 (image classification)~\cite{he2016deep},
because currently TPU only supports TensorFlow.
We also select other real-world deep learning workloads~\cite{tpurepo}, including
RetinaNet~\cite{lin2017focal},
DenseNet~\cite{huang2017densely}, MobileNet~\cite{howard2017mobilenets},
and SqueezeNet~\cite{iandola2016squeezenet}.
We refer to them as real workloads or real models.
The batch sizes are the largest supported on the hardware platform.
For example, on TPU with bfloat16, we use batch size 64 for RetinaNet, 4k for Transformer, and 1024 for the rest of the workloads.

\begin{figure}[t]
\begin{center}
\includegraphics[width=0.8\columnwidth]{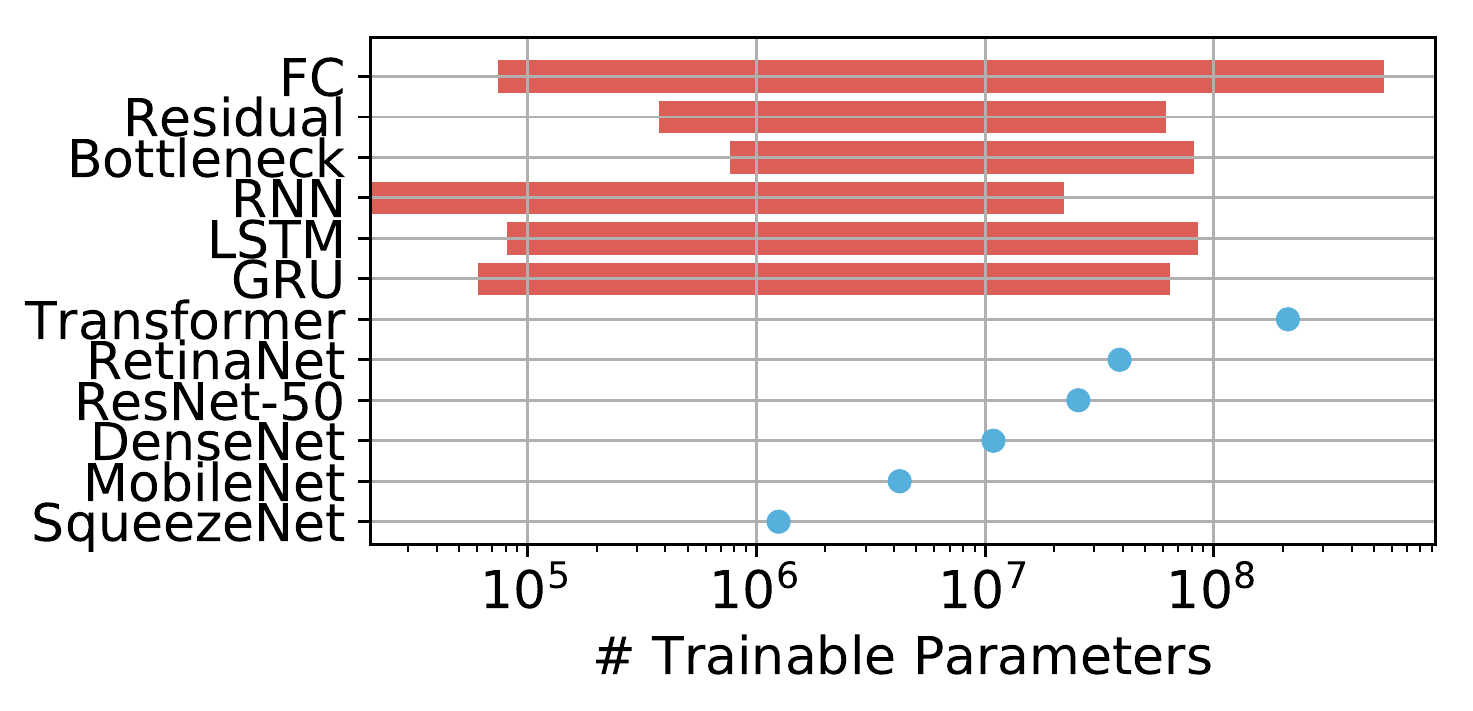}
\vspace{-1em}
\caption{The numbers of trainable parameters for all workloads in this paper. Those from ParaDnn range from 10k to nearly a billion parameters, which is larger the range of real workload sizes, shown as dots.}
\vspace{-2em}
\label{fig:model_params}
\end{center}
\end{figure}

\Fig{fig:model_params} shows the numbers of trainable parameters across all workloads to quantify the size of the models.
The ParaDnn workloads are shown as ranges and the real workloads as dots.
ParaDnn covers a large range of models, from 10k to nearly a billion parameters.
Transformer is the largest real FC, and RetinaNet is the largest real CNN.
The small models, SqueezeNet and MobileNet, reflect models typically targeted towards 
mobile applications. RetinaNet and ResNet-50 provide state-of-the-art image classification 
accuracy.

\section{Hardware Platforms}
\label{sec:platform}
Our selection of hardware reflects the latest configurations widely available in cloud platforms at paper submission time.
Platform specifications are summarized in Table~\ref{table:platforms}.

\paragraph{CPU Platform} The CPU is an n1-standard-32 instance from Google Cloud Platform with Skylake architecture.
It has 16 cores and 32 threads.
It has the largest memory (120 GB) and lowest peak flops (2 TFLOPS) among the three.
GeekBench~4 produced the bandwidth measurement.

\paragraph{GPU Platform} The GPU is an NVIDIA V100 in a DGX-1 GPU platform that contains 8 V100 packages (SXM2) connected via 300 GB/s NVlink 2.0 interconnect.
We currently measure the performance of a single SXM2 node. 
One node has 16 GB of memory and 900 GB/s memory bandwidth.
A V100 has 640 tensor cores and is able to run mixed precision training using float16 to compute and float32 to accumulate, making its peak performance 125 TFLOPS.

\paragraph{TPU Platform} The TPU is a Cloud TPU instance to which we were given academic access in February 2018.
Its system architecture includes a Cloud Engine VM, a Cloud TPU server, Google Cloud storage, and a Cloud TPU board~\cite{cloudtpu}.
Each TPU board contains four TPU packages (the default Cloud TPU configuration)~\cite{dean2017hotchips}.
One TPU v2 package supports 45 TFLOPS and contains 2 cores.
One core has one matrix unit (MXU).
Total ML acceleration for a Cloud TPU v2 platform is \SI{180}{TFLOPS}.
Memory size is \SI{8}{GB} per core, or \SI{64}{GB} per board, with \SI{2400}{GB/s} overall memory bandwidth.
TPU v2 supports mixed precision training, using bfloat16 to compute and float32 to accumulate.
Compared to v2, TPU v3 doubles the number of MXUs and HMB capacity per core~\cite{cloudtpu}.
The memory bandwidth has not been disclosed, but empirical results show that it is increased by 1.5$\times$.
TPU v3 has a peak of \SI{420}{TFLOPS}, 2.3$\times$ greater than v2, likely because of higher frequency.
Because v3 is an upgrade from v2, we focus on studying v2.
In this paper, TPU refers to Cloud TPU v2, unless specified otherwise.

{\it Understanding TPU memory size.}
Data parallelism is implemented on the TPU, where one batch of training data is split evenly and sent to the 8 cores on the TPU board.
The model is not distributed; every TPU core keeps a whole copy of it.
Therefore
memory size per core determines the maximum model supported,
while total on-board memory determines the maximum data batch size.
That is why in \Sec{sec:xcompare:fc}, the GPU platform supports larger models than the TPU, and the TPU supports larger batch sizes (\Sec{sec:xcompare:cnnrnn}).

{\it Comparison rationale.}
We evaluate one V100 package and one TPU board (4 packages) because they are the minimal units available.
The configurations are encapsulated.
On Cloud TPU, distribution of computation across the four TPU packages on a TPU board happens automatically.
On the other hand, multi-GPU performance depends largely on the user's implementation. Multi-GPU/TPU performance
is beyond the scope of this work as discussed in \Sec{sec:limitation}.
Therefore, note that conclusions in this paper do not apply to multi-GPU or larger TPU systems.





\begin{table}[t]
\vspace{-1em}
\small
\begin{center}
\begin{threeparttable}
\begin{tabular}{|c|c|c|c|c|c|c|}

\multicolumn{1}{c}{} & \multicolumn{1}{c}{} &\multicolumn{1}{c}{} &\multicolumn{1}{c}{\emph{Mem}} & \multicolumn{1}{c}{\emph{Mem}} & \multicolumn{1}{c}{\emph{Mem Bdw}} & \multicolumn{1}{c}{Peak} \\ 

\multicolumn{1}{c}{\emph{Platform}}  & \multicolumn{1}{c}{\emph{Unit}}  & \multicolumn{1}{c}{\emph{Version}} & 
 \multicolumn{1}{c}{\emph{Type}} & \multicolumn{1}{c}{\emph{(GB)}} &\multicolumn{1}{c}{\emph{(GB/s)}} & \multicolumn{1}{c}{\emph{FLOPS}} \\
\hline 

CPU & 1 VM & Skylake & DDR4 & 120 & 16.6 & 2T SP\tnote{$\dagger$}\\
\hline 

GPU & 1 & V100 &  &  &  &  \\

(DGX-1) &  Pkg & (SXM2) & HBM2 & 16 & 900 & 125T \\

\hline 

  & 1 Board &   &  &  &   & \\
 
 TPU & (8 cores) & v2 & HBM & 8 & 2400 & 180T \\
\hline 

 TPUv3 & 8 cores & v3 & HBM & 16 & 3600\tnote{*} & 420T \\
\hline 
\end{tabular}
\begin{tablenotes}
\item[$\dagger$]Single precision: 2 FMA $\times$ 32 SP $\times$ 16 cores $\times$ 2G frequency = 2 SP TFLOPS
\item[*]Estimated based on empirical results (\Sec{sec:tpu:v3}).
\end{tablenotes}
\end{threeparttable}
\end{center}
\vspace{-1.5em}
\caption{Hardware platforms under study.}
\vspace{-1.5em}
\label{table:platforms}
\end{table}

\section{TPU Architectural Implications}
\label{sec:tpu}

As the end of Dennard scaling and Moore's law has slowed the performance improvement of general-purpose microprocessors~\cite{hennessynew},
the design of domain-specific hardware is becoming more and more relevant.
The TPU is a prominent example of domain-specific hardware~\cite{jouppi2017datacenter,dean2017hotchips}.
Its development was motivated by the observation that, with conventional CPUs, Google would have had to double their datacenter footprint to meet the internal demand for machine learning workloads.
Google has been using TPUs for their large-scale production systems, including Search, Translate, and Gmail.
Analyzing the architecture of such systems can provide valuable insights into future deep learning accelerator design.

In this section, we study the performance characteristics of TPU v2 and v3~\cite{dean2017hotchips,cloudtpu} with a focus on v2,
from the computation capability in the core (FLOPS) to the system balance.
Based on our observations, we discuss possible steps to improve TPU performance, which can be generalized to other deep 
learning accelerator systems. The following is a summary of our key observations and insights:

\begin{itemize}
\vspace{-0.5em}
\item \textbf{FLOPS (\Sec{sec:tpu:flops})}: TPU makes good use of the parallelism exposed by batch size and model width, but parallelism due to model depth is under-exploited, suggesting opportunities for model pipelining~\cite{gpipe}.

\vspace{-0.5em}
\item \textbf{Memory bandwidth (\Sec{sec:tpu:roofline})}: Memory bandwidth is the performance bottleneck of many models. Even highly-optimized compute-bound models show a significant fraction of memory-bound operations (13\% in ResNet-50).
    Improving memory access for such operations is key to further performance improvement.

\vspace{-0.5em}
\item \textbf{Multi-chip overhead (\Sec{sec:tpu:multinode})}: Communication overhead in a multi-chip system is non-negligible (up to 13\% for CNNs with sizes similar to ResNet-50) but can be amortized with large batch sizes.
    Reducing the communication overhead can lead to performance gain.
    
\vspace{-0.5em}
\item \textbf{Host-device balance (\Sec{sec:tpu:systembalance})}: Data quantization can make compute-bound workloads data-infeed-bound.
    Resolving the data-infeed bottleneck can improve performance by at least 34\%.
    
\vspace{-0.5em}
\item \textbf{TPU v3 (\Sec{sec:tpu:v3})}: The maximum speedup of TPU v3 over v2 is up to 3$\times$, exceeding the 2.3$\times$ FLOPS increase.
TPU v3 benefited from its doubled memory capacity (which allows twice the batch size of v2) as well as increased memory bandwidth.
    
\end{itemize}

\subsection{FLOPS Utilization}
\label{sec:tpu:flops}

Floating point operations per second (FLOPS) utilization is the ratio of average FLOPS to peak FLOPS, measuring how efficiently the computation capacity of a platform is used.
We discuss the TPU FLOPS utilization of the parameterized models in this section.
We first visualize how the model hyperparameters listed in \Tbl{table:var_range} affect FLOPS utilization.
Then we introduce an analysis methodology to quantify the hyperparameter effect using linear regression.

\begin{figure}[t]
    \centering
        \subfloat[FC]{\includegraphics[width=0.35\columnwidth]{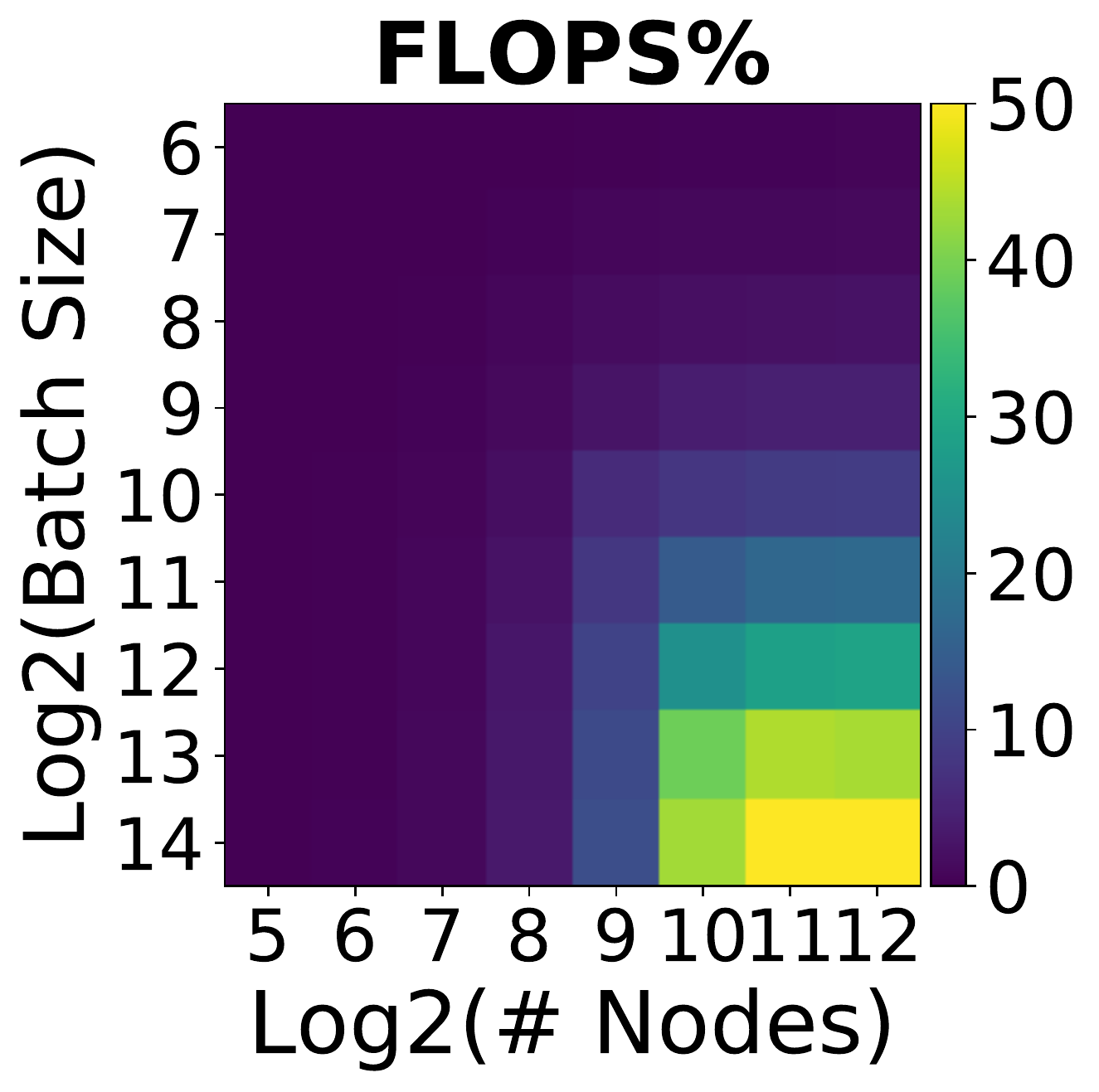}}
        \subfloat[CNN]{\includegraphics[width=0.29\columnwidth]{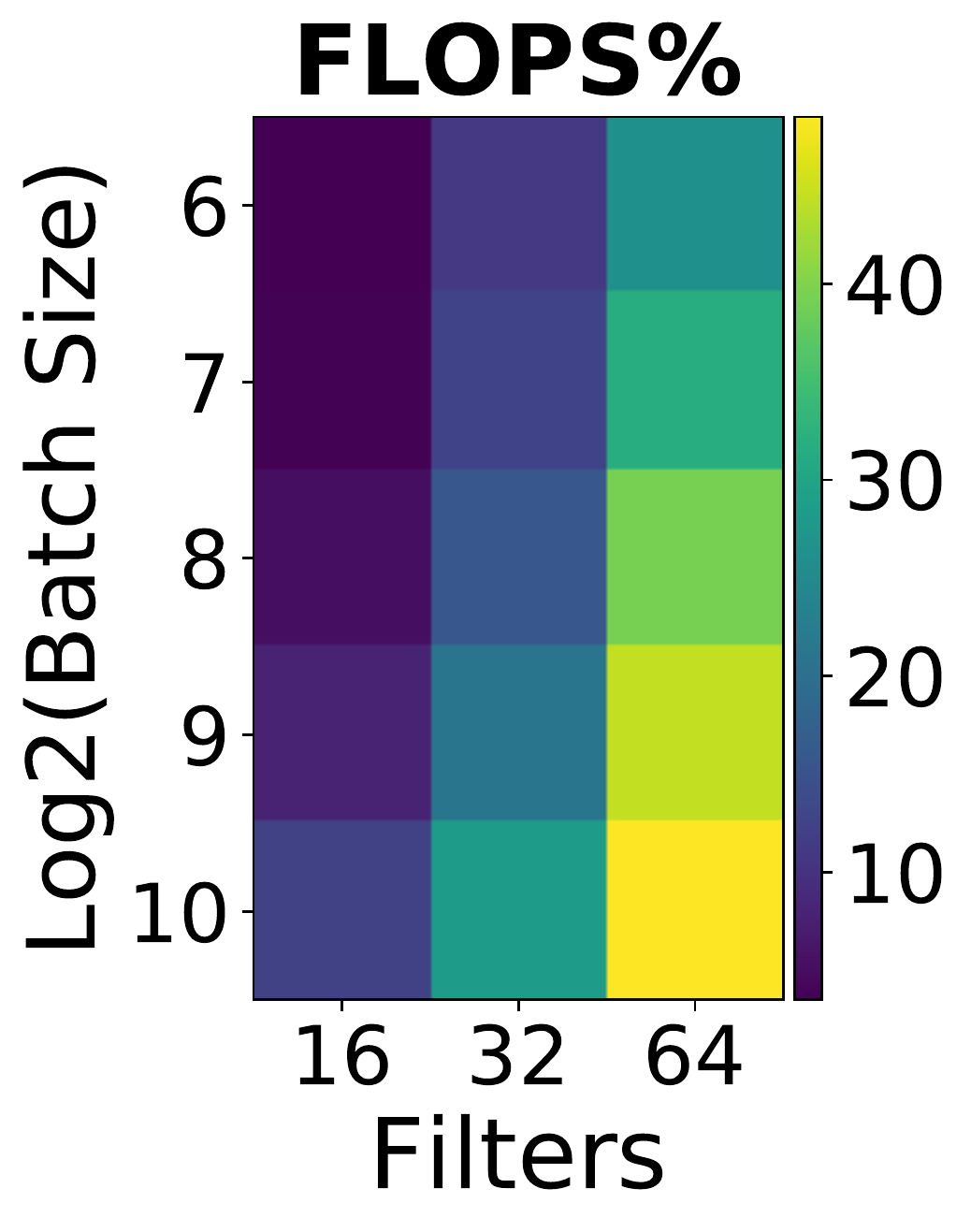}}
        \subfloat[RNN]{\includegraphics[width=0.35\columnwidth]{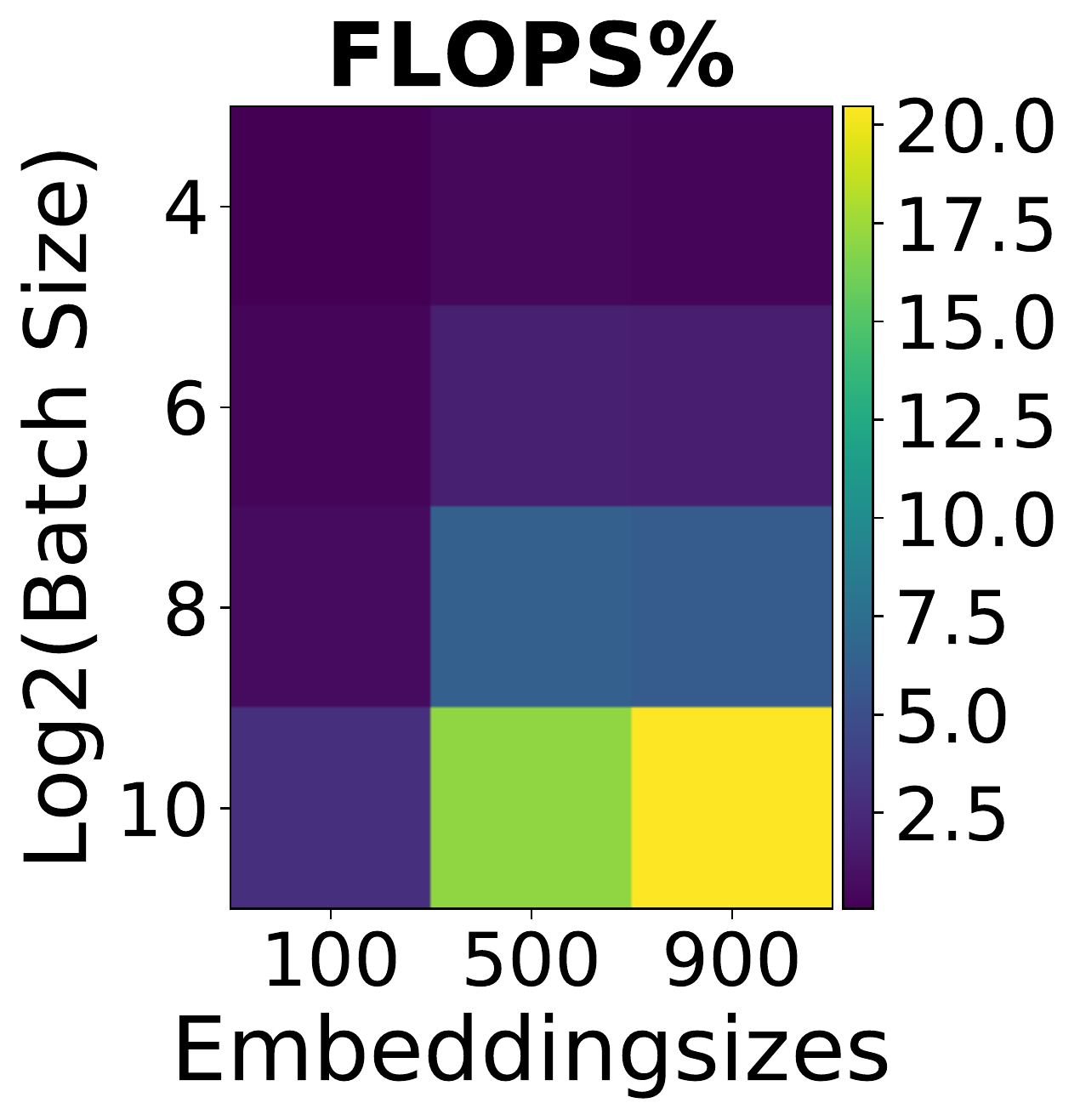}}
        \qquad
       \subfloat[FC]{\includegraphics[width=0.31\columnwidth]{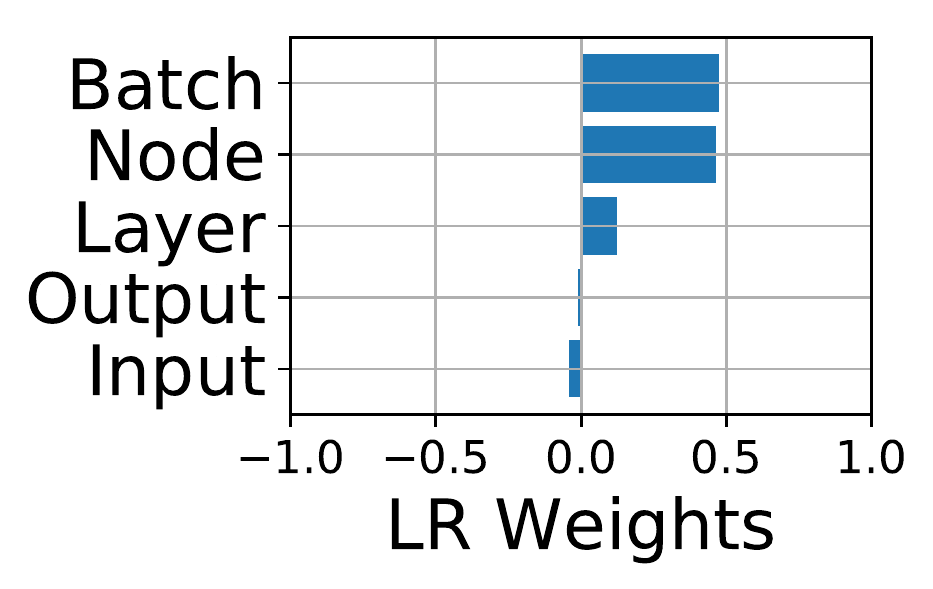}}
        \subfloat[CNN]{\includegraphics[width=0.31\columnwidth]{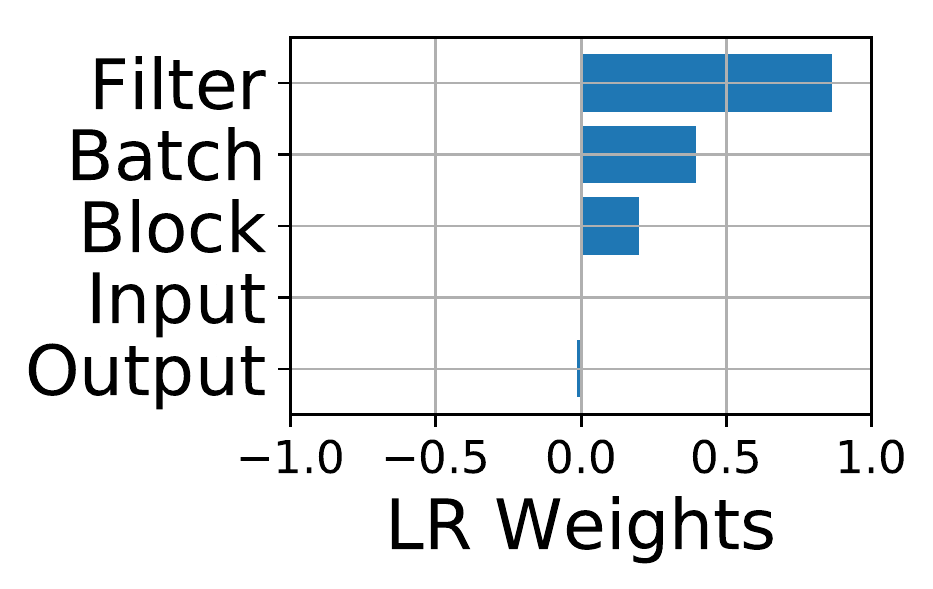}}
        \subfloat[RNN]{\includegraphics[width=0.36\columnwidth]{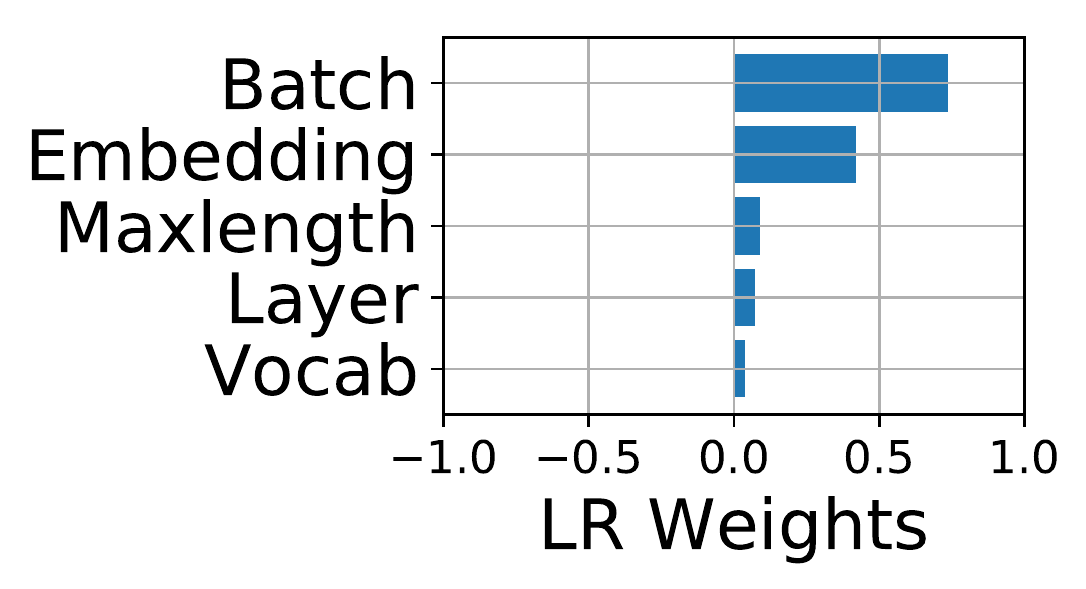}}
        \vspace{-1em}
    \caption{FLOPS utilization and its correlation with hyperparameters. (a)--(c) show FLOPS utilization of parameterized models. (d)--(f) quantify effects of model hyperparameters on FLOPS utilization, using linear regression weights.}
    \vspace{-1em}
    \label{fig:tpu_flops}
\end{figure}

\paragraph{FLOPS Utilization Heat Maps}
Figure~\ref{fig:tpu_flops}(a)--(c) presents heat maps of FLOPS utilization for FC, CNN, and RNN models, obtained by sweeping the hyperparameters with ranges listed in \Tbl{table:var_range}.
We choose two hyperparameters for each model type that affect FLOPS utilization the most (see below for how we choose them) and show them on the $x$- and $y$-axes while keeping the other hyperparameters fixed.
Specifically, we fix layer (32), input (2000), and output units (1000) for FCs,
block (6), input image size ($300\times300\times3$), and output unit (1000) for CNNs, and
layer (9), vocabulary size (32), and max length (50) for RNNs.

Figures~\ref{fig:tpu_flops}(a)--(c) show that the FLOPS utilization of all three models increases with batch size.
Other than that, the FLOPS utilization of FCs increases with number of nodes per layer (Figure~\ref{fig:tpu_flops}(a)), that of CNNs increases with filters, and that of RNNs with embedding size.
This indicates that TPU is capable of leveraging the parallelism within a batch (the former) and within the width of the models (the latter).

\paragraph{Studying Parameterized Models with Linear Regression}
Having discussed the qualitative effects of hyperparameters on FLOPS utilization, we now build a linear regression (LR) model and use the weights to quantify these effects.
Note that the LR model is only for measuring the effects of hyperparameters.
We do not use it for prediction.

In the case of FC, the linear regression model is
\begin{align*}
	\mbox{FLOPS} =~&w_0 \times \mbox{layer} + w_1 \times \mbox{node}~+ \\[-.75ex]
                       &w_2 \times \mbox{input} + w_3 \times \mbox{output} + w_4 \times \mbox{batch size},
\end{align*}
where $w_0$--$w_4$ are the weights of the hyperparameters.
To train the LR model, all the values are normalized to the same scale,
so that we can use the weights as a measure of importance.
For example, positive $w_1$ indicates that node count affects performance positively.
If the absolute value of $w_1$ is larger than that of $w_0$, it indicates node count has a larger effect on FLOPS than layer count.
Other similar metrics for feature selection, including T-test and F-test, may be used for this purpose~\cite{hogg2005introduction}.
We choose LR mainly to get the signs of the weights, which indicate the positive or negative effects of the hyperparameters on performance, while T-test and F-test only report positive values as importance.

Figures~\ref{fig:tpu_flops}(d)--(f) show the LR weights of the model hyperparameters.
The $x$- and $y$-axes in Figures~\ref{fig:tpu_flops}(a)--(c) are the hyperparameters with the highest absolute values in Figures~\ref{fig:tpu_flops}(d)--(f).
\Fig{fig:tpu_flops}(d) shows that the FLOPS utilization of FC is largely affected by batch size and node, while layer, output, and input do not matter as much.
Similarly, \Fig{fig:tpu_flops}(e) shows filter is the most important, and batch size is more important than block, while input and output have minimal impact.
The TPU FLOPS of RNNs is not affected by maximum length, number of layers, or vocabulary size.

\paragraph{Architectural Implications}
The TPU takes advantage of parallelism due to large batch size
and model width, including that from nodes per layer in FC, filters in CNN, and embedding sizes in RNN.
Parallelism opportunities from large numbers of layers remain to be explored, by approaches such as model parallelism~\cite{dean2012large,jia2018beyond} and pipelining~\cite{gpipe}.

\subsection{Roofline Model Analysis}
\label{sec:tpu:roofline}

The FLOPS utilization in the previous section shows the computation capability of TPU, but the core is only part of the problem when designing an accelerator.
In particular, memory bandwidth is another important aspect that can have significant impact on performance.
In this section, we use the roofline model~\cite{williams2009roofline} to analyze the computation and memory bandwidth of FCs and CNNs.
Roofline models are useful to demonstrate memory and computation bottlenecks~\cite{williams2009roofline,jouppi2017datacenter}.
We omit RNN models because the TPU profiler reports incorrect numbers for memory bandwidth of RNN models.



\begin{figure}[t]
    \centering
        \subfloat[FC bfloat16]{\includegraphics[width=0.35\columnwidth]{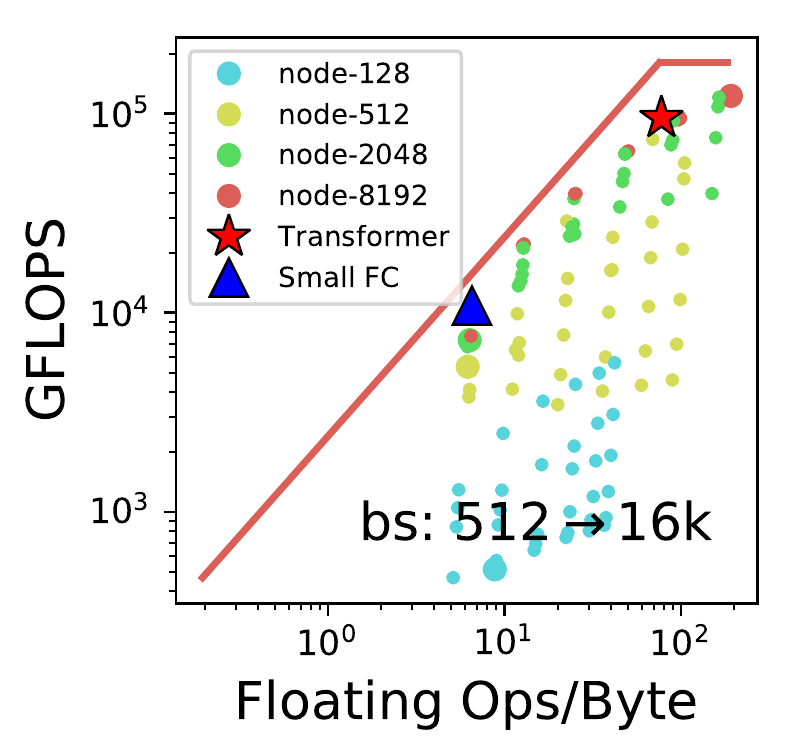}}
        \subfloat[FC Op Breakdown]{\includegraphics[width=0.65\columnwidth]{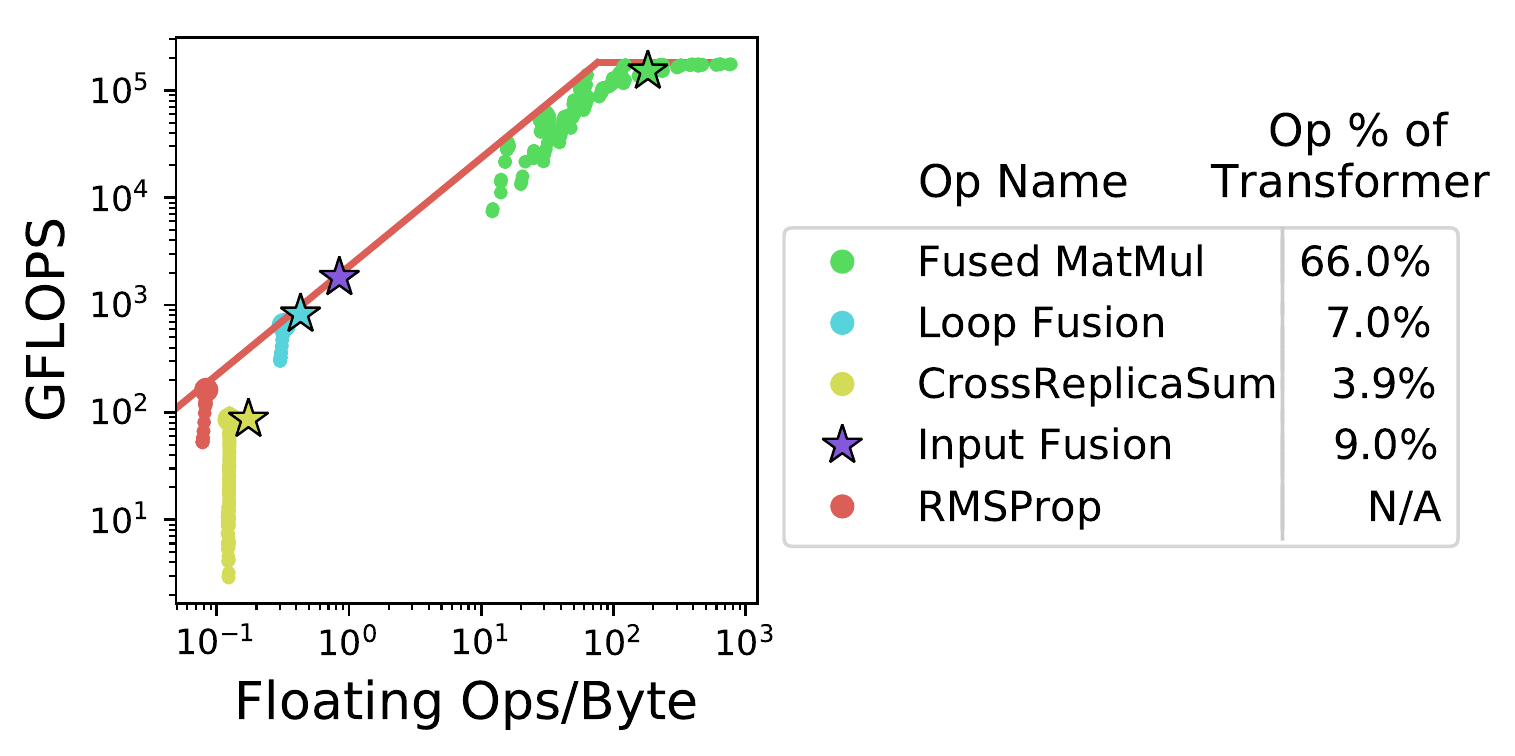}}
        \qquad
       \subfloat[CNN bfloat16]{\includegraphics[width=0.35\columnwidth]{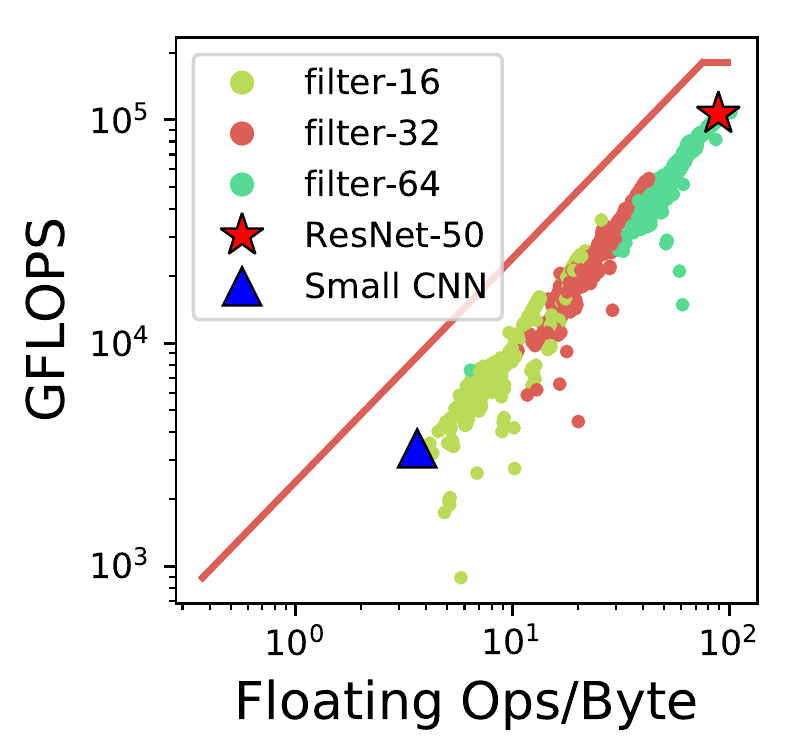}}
        \subfloat[CNN Op Breakdown]{\includegraphics[width=0.65\columnwidth]{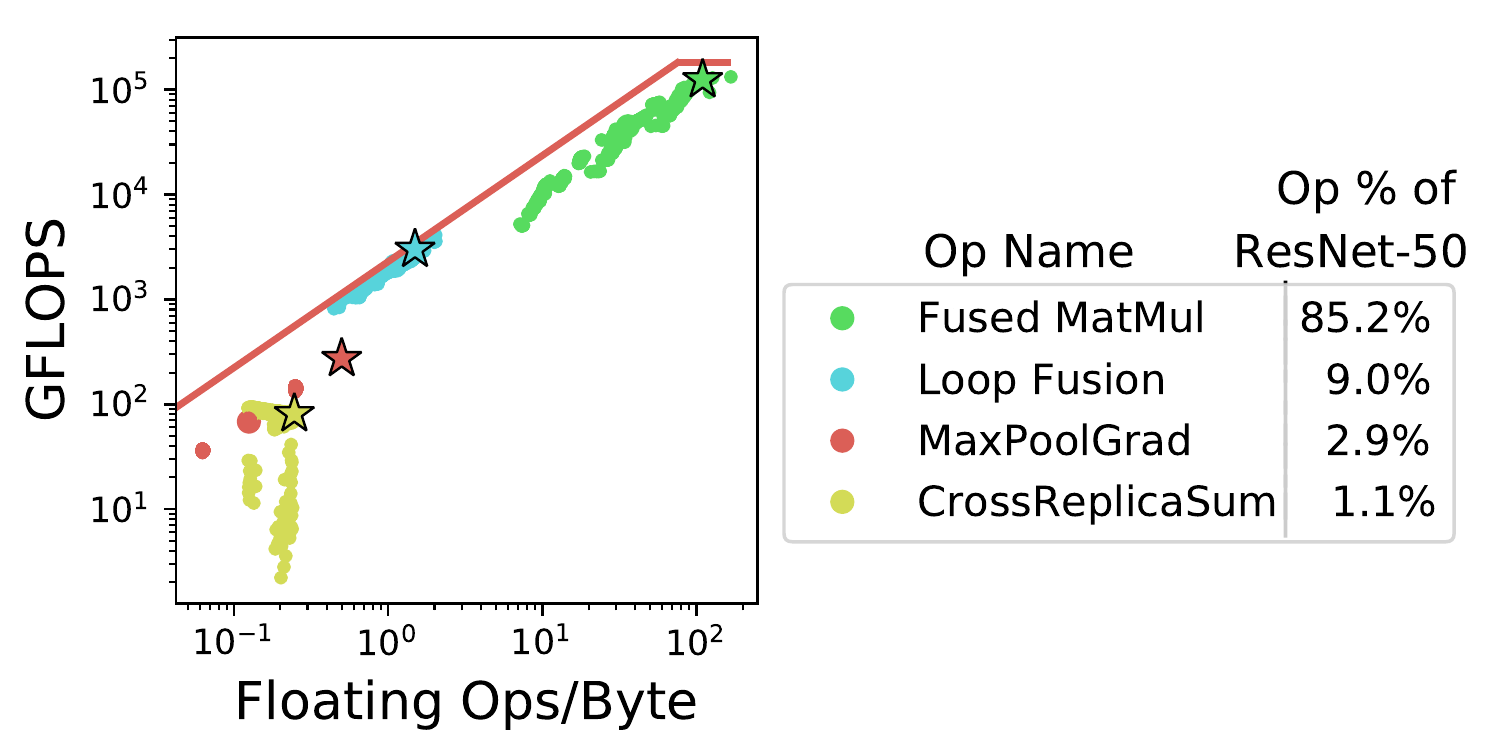}}
        \vspace{-0.5em}
    \caption{Rooflines for FC and CNN on TPU. Workloads with matrix multiply (MatMul) operations are compute-bound. Even compute-bound workloads like Transformer and ResNet-50 have more than 10\% memory-bound operations. (a) and (c) show rooflines of parameterized and real-world models. (b) and (d) show the operation breakdown.}
    \vspace{-1.5em}
    \label{fig:tpu_rooflines}
\end{figure}

\paragraph{The Roofline Model}
\Fig{fig:tpu_rooflines} shows the roofline plots.
The $y$-axis is FLOPS and the $x$-axis is arithmetic intensity, i.e., floating-point operations per byte transferred from memory.
The roofline (the red line in \Fig{fig:tpu_rooflines}) has of a slanted part and a horizontal part.
It represents the highest achievable FLOPS at a given arithmetic intensity.
Any data point $(x,y)$ on the slanted part has $x/y=\mbox{memory}~\mbox{bandwidth}$.
The horizontal part is the peak FLOPS on the hardware.
A workload or operation (a point in \Fig{fig:tpu_rooflines}) close to the slanted roofline is memory-bound; one close to the horizontal part is compute-bound. 
A workload or operation not close to the roofline stresses neither memory interconnect nor compute units.

Figures~\ref{fig:tpu_rooflines}(a) and \ref{fig:tpu_rooflines}(c) show all the parameterized FC and CNN models (dots) plus Transformer and ResNet-50 (stars).
Figures~\ref{fig:tpu_rooflines}(b) and \ref{fig:tpu_rooflines}(d) show all the operation breakdowns.
Transformer and ResNet-50 are just instances (sparse design points) in ParaDnn, so the stars overlap some of the dots.
This is because ParaDnn enables more comprehensive model architecture design space exploration and supports benchmarking hardware systems more systematically.
An exception is that some operations of Transformer do not align closely with those of FCs. This results from a choice in this paper, not a fundamental flaw of ParaDnn.
ParaDnn uses the RMSProp optimizer, keeping nodes per layer uniform in a parameterized FC,
while Transformer uses the \textit{adafactor} optimizer and 
has layers with 4k, 2k, and 512 nodes.

\paragraph{FC}
\Fig{fig:tpu_rooflines}(a) shows that large batch sizes make FCs more compute-bound, and more nodes make FCs more memory-bound.
That is because FCs with more nodes need to transfer more weights/activations from the memory,
and large batch sizes increase the computation per weight/activation transferred, i.e, the arithmetic intensity.
For example, for FCs with $\ge \mbox{2k}$ nodes, using large batch sizes turns memory-bound FCs into compute-bound.
Specifically, the FCs with $\ge \mbox{2k}$ nodes per layer and $\ge \mbox{8k}$ batch size are compute-bound.
Transformer is close to compute-bound and it uses 4k batch size, which causes it to overlap with FCs having 4k batch sizes.

\paragraph{CNN}
\Fig{fig:tpu_rooflines}(c) shows that models close to ResNet-50 are compute-bound, while a majority of the CNNs are bottlenecked by memory bandwidth.
As it is in log scale, it shows that practically achievable memory bandwidth for the CNNs is less than the theoretical bandwidth.
The CNNs' higher FLOPS comes from higher arithmetic intensity caused by more filters.
When memory bandwidth is the bottleneck, the way to increase FLOPS is to increase arithmetic intensity.

\paragraph{Operation Breakdown}
The triangles in
Figures~\ref{fig:tpu_rooflines}(a) and \ref{fig:tpu_rooflines}(c) are selected memory-bound models. The FC has 8 layers, 8192 nodes per layer, and batch size 512; the CNN has 1 block per group, 16 filters, and batch size 64.
Figures~\ref{fig:tpu_rooflines}(b) and \ref{fig:tpu_rooflines}(d) show the TensorFlow operations taking more than 1\% of the workload execution time and more than 0 TPU FLOPS.
The arithmetic intensity of such operations can be as low as $0.125$.\footnote{For example, an activation accumulation operation (CrossReplicaSum in TensorFlow) uses float32 even with bfloat16 model weights.
In this case, the arithmetic intensity is $1 / (2 \times 4~\mbox{bytes}) = 0.125$, i.e., one floating point addition for every two data points loaded.}
The TensorFlow breakdown in \Fig{fig:tpu_rooflines} is generated after operation fusion, which is a technique combining and executing several operations together for higher efficiency.

\paragraph{Large MatMuls}
Figures~\ref{fig:tpu_rooflines}(b) and \ref{fig:tpu_rooflines}(d) show that the only compute-bound operation is large fused MatMul (matrix multiply fused with other operations), so a compute-bound model needs to have compute-bound MatMuls.
Other operations are closer to the slanted line, indicating they are constrained by memory bandwidth.
For example, in \Fig{fig:tpu_rooflines}(a) and (c), Transformer and ResNet-50 are compute-bound
because they have compute-bound MatMuls in Figures~\ref{fig:tpu_rooflines}(b) and \ref{fig:tpu_rooflines}(d).

\paragraph{Memory-bound Operations}
Interestingly, even compute-bound FC/CNN models contain a noticeable fraction of memory-bound operations. 
Transformer has three memory-bound operations: (1)~input fusion (9.0\%), which includes multiply, subtract, and reduce; 
(2)~loop fusion (7.0\%), which consists of control flow operations (e.g., select and equal-to); and (3)~CrossReplicaSum (3.9\%), which sums up the values across multiple weight replicas. 
These three operations contribute to 19.9\% of the total execution time.
(12.3\% of the execution time is for data formatting, which has no arithmetic intensity or TPU FLOPS.)
Even compute-bound ResNet-50 has many memory-bound operations, including 
loop fusion (9\%), MaxPoolGrad (2.9\%), and CrossReplicaSum (1.1\%), which sums to 13\%, showing the need for both end-to-end and per-operation optimization for deep learning accelerators.


\paragraph{Architectural Implications}
Compute-bound FCs and CNNs have large MatMul operations.
Surprisingly, even compute-bound models contain non-negligible fractions (19.9\% for Transformer and 13\% for ResNet-50) of memory-bound operations.
Given the current TPU system, memory-bound operations need more attention.
Potential ways to speed up memory-bound operations include increasing memory bandwidth and reducing memory traffic.
Traditional architectural efforts to reduce memory traffic can be adopted, such as exploiting the memory locality by caching~\cite{hennessy2011computer}.
Software/compiler approaches include better operation fusion~\cite{xlajit,chen2018tvm,rotem2018glow}, more aggressive data quantization~\cite{banner2018scalable}, and weights and gradients compression~\cite{han2015deep,lin2017deep}.

\subsection{Multi-Chip Overhead}
\label{sec:tpu:multinode}

This section analyzes communication overhead in a multi-chip system.
Previous sections focus on the compute and memory bandwidth of a TPU core.
But these are not the only factors that affect training performance, because typical large-scale training systems use multiple chips~\cite{dean2012large}.
This section evaluates the scalability of a multi-chip TPU system.

\begin{figure}[t]
    \centering
        \subfloat[FC]{\includegraphics[width=0.4\columnwidth]{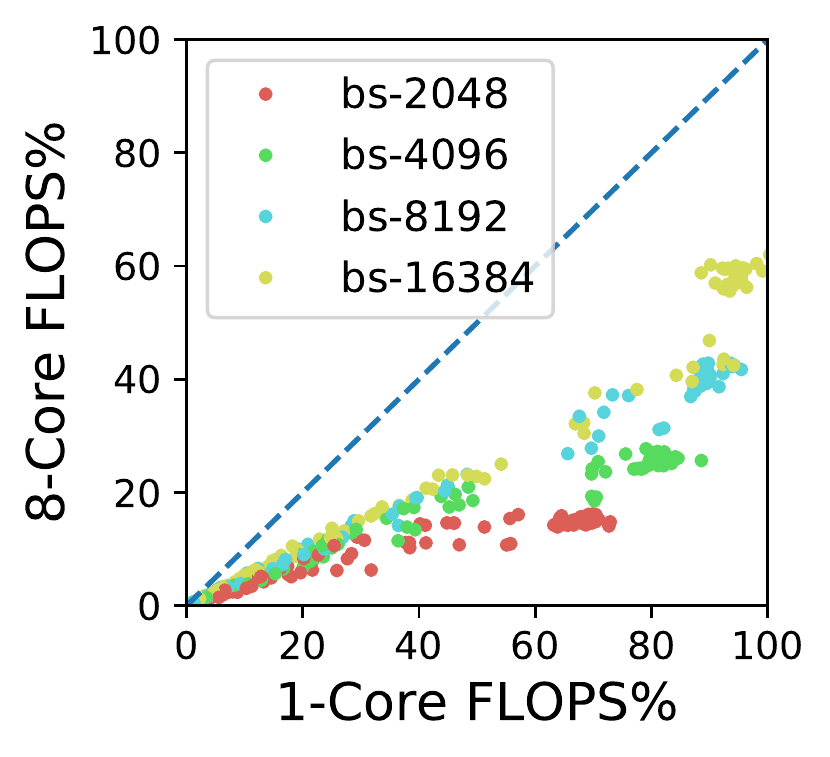}}
        \subfloat[CNN]{\includegraphics[width=0.4\columnwidth]{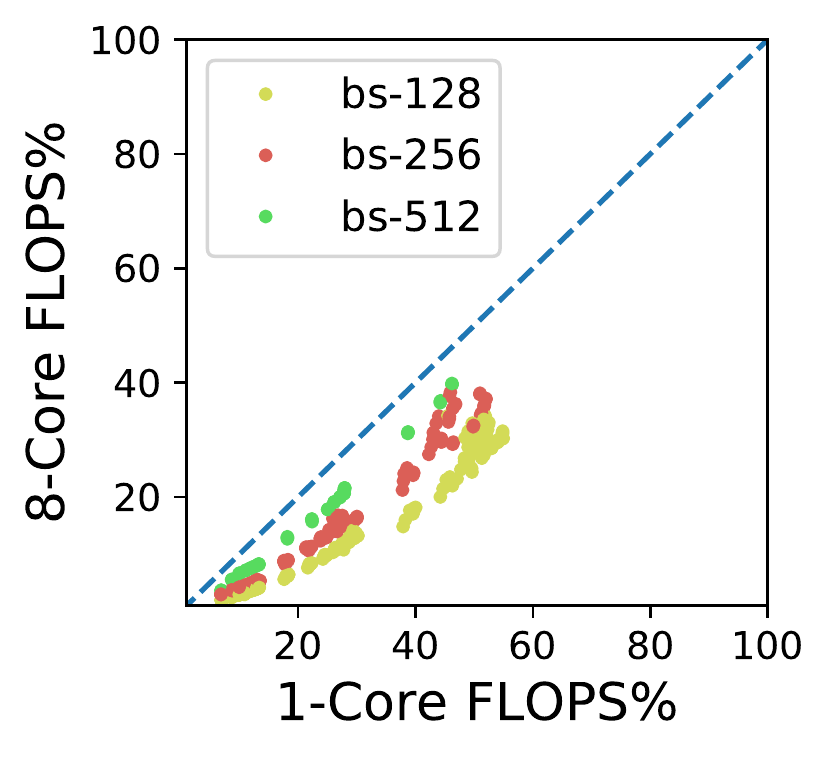}}
        \vspace{-1em}
    \caption{Communication overhead in a multi-chip system is non-negligible, but is reduced with large batch sizes.}
    \vspace{-1.5em}
    \label{fig:tpu_multinode}
\end{figure}

To quantify the multi-chip overhead, we compare the FLOPS utilization of 1-core ($x$-axis) and 8-core TPU ($y$-axis) in 
\Fig{fig:tpu_multinode}.
If there were no multi-chip overhead, FLOPS utilization of 1-core and 8-core should be the same, i.e., 
all points should lie on the dashed line in \Fig{fig:tpu_multinode} showing $x=y$.
On the 8-core TPU, FCs need at least 16k batch size to achieve more than 50\% FLOPS utilization.
Specifically, FCs with $\ge 256$ nodes and $\le 512$ batch size are faster to run on 1-core TPU than on 8-core TPU.
Therefore we consider FCs with larger than 1024 batch size in \Fig{fig:tpu_multinode}.

As shown in the figure, 8-core TPU shows noticeably lower FLOPS utilization than 1-core TPU, indicating significant 
inter-core communication overhead.
For FC, the maximum FLOPS utilization in 8-core TPU is 62\%, compared to 100\% in 1-core TPU. 
Multi-chip overhead is less noticeable in CNNs, with FLOPS utilization decreasing 
from 55\% in 1-core TPU to 40\% in 8-core. It is worse for FCs because there
are more weights to synchronize across the TPU cores than for CNNs. Based on Amdahl's law, we 
calculate that the maximum non-parallel fraction of the workloads is 
up to 60\% for FC and 40\% for CNN. The FLOPS utilization difference 
is smaller with larger batch sizes for both FC and CNN,
because it increases the computation without increasing the weight synchronization.
Using the largest batch size shown in \Fig{fig:tpu_multinode}, 
the 90th-percentile of non-parallel fractions are 16\% for FC and 8.8\% for CNN.

\paragraph{Architectural Implications}
We show that communication overhead in multi-chip systems is non-negligible even for large FCs and CNNs.
Using large batch size can reduce the overhead by
increasing the computation parallelism without increasing weight transfers.
Possible optimizations include
relaxed synchronization, model parallelism~\cite{dean2012large}, 
gradient compression~\cite{lin2017deep},
and algorithm and architecture support for weight pruning and compression~\cite{han2015deep} before synchronization.

\subsection{Host-Device Balance}
\label{sec:tpu:systembalance}

Previous subsections have focused on the performance of the accelerator itself. This section focuses 
on ``data infeed,'' the process of preparing and moving input data to the TPU board.
ParaDnn analysis avoids part of the data infeed overhead by synthesizing data on the CPU host.
We now describe a case study with real-world workloads to show the importance of balancing accelerators and the host in a system.

\begin{figure}[t]
\begin{center}
\includegraphics[width=.9\columnwidth]{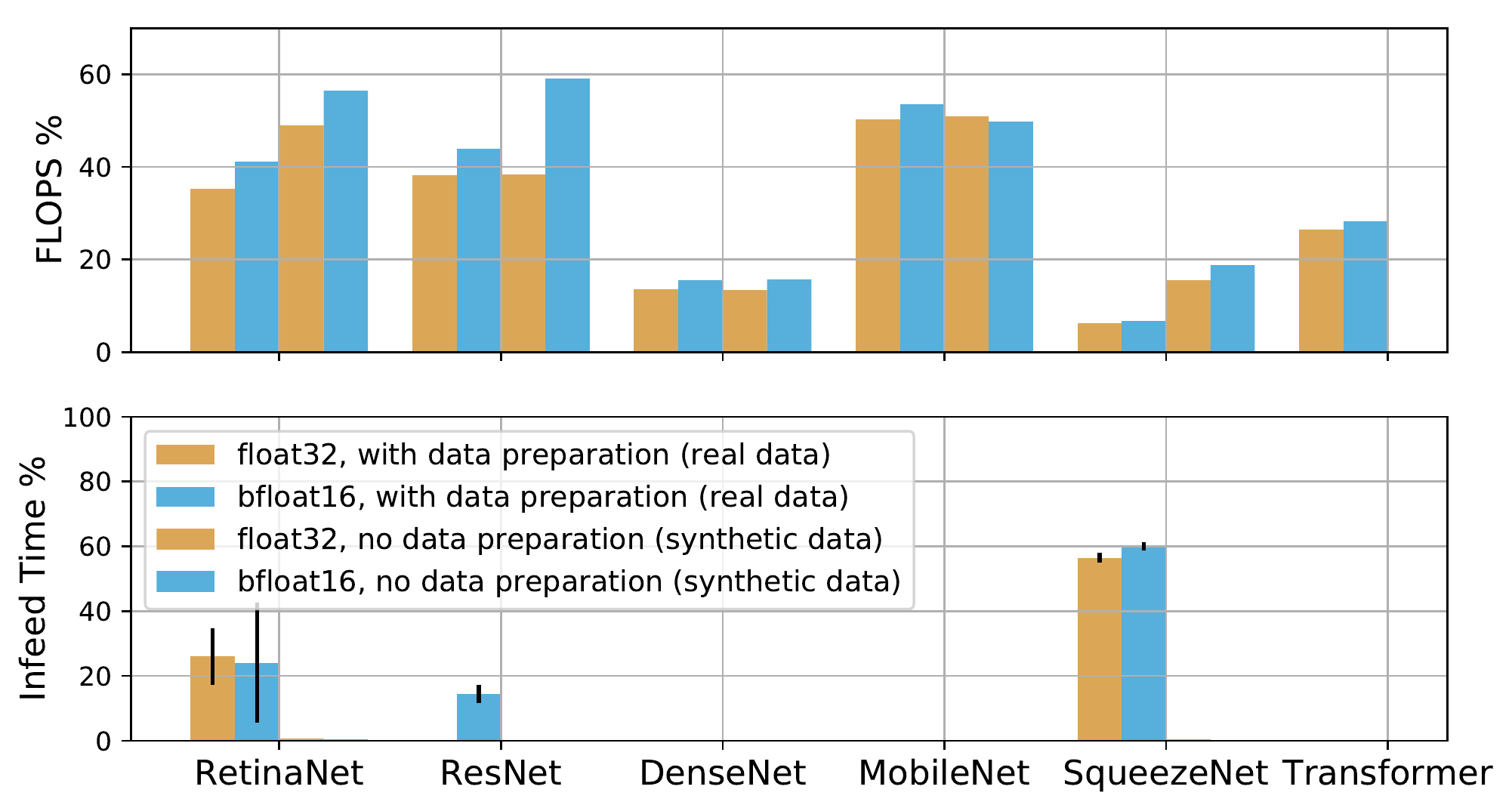}
\vspace{-0.5em}
\caption{FLOPS utilization (top) and infeed time (bottom) of the real models using float32 and bfloat16, with and without data preparation. Models with large infeed time percentage, i.e., RetinaNet and SqueezeNet, are limited by data infeed.}
\vspace{-2em}
\label{fig:casestudy}
\end{center}
\end{figure}

\paragraph{TPU Device and Host}
The TPU system is composed of a CPU host and a TPU device~\cite{dean2017hotchips}.
For real-world CNNs, the host fetches images from the network, decodes, 
preprocesses, and feeds them to the device. \Fig{fig:casestudy} calls this data preparation.
The device then performs training computation on the images.
Data infeed means
network overhead, host compute, and bandwidth 
between host and device.

\paragraph{Infeed Overhead Analysis}
To quantify the infeed overhead, we run real-world workloads both with and without data preparation, by directly 
feeding synthetic data as postprocessed inputs. We also compare models using float32 to those with bfloat16, because replacing float32 with bfloat16 can affect the execution time of both data infeed and device computation.
First, the arithmetic intensity of all operations doubles, because the same computation can be performed with half of the bytes transferred.
Second, the FLOPS of memory-bound operations improves in the device, because increased arithmetic intensity moves those operations towards the upper right in the roofline model of \Fig{fig:tpu_rooflines}.
Third, improved device performance increases the need for faster data infeeding, which puts more pressure on the host.

\Fig{fig:casestudy} shows FLOPS utilization and infeed time of the real-world workloads.
FLOPS utilization measures computation efficiency and infeed time measures how long the device waits for data, both of which are collected from the TPU profiler.
The error bars are one standard deviation of the one-minute samples from the profiler.

The figure shows that the bottleneck of a workload can be on the device or in data infeed by different degrees under different circumstances.
Data infeed bottlenecks RetinaNet and SqueezeNet, as the performance increases noticeably when data preparation is skipped.
Eliminating that bottleneck brings 37\% and 180\% speedup, respectively, for RetinaNet and SqueezeNet using bfloat16.
RetinaNet's bottleneck is likely because it uses the COCO dataset ($640 \times 640$ images), while others use the ImageNet dataset ($224 \times 224$ images).

ResNet-50 is bottlenecked by the device when using float32, and by data infeed when using bfloat16.
That bitwidth reduction speeds device execution and increases FLOPS utilization so that training throughput on the device surpasses data preparation throughput on the host.
If the resulting data infeed bottleneck can be resolved, the performance of bfloat16 ResNet-50 can be improved by 34\%.
Switching RetinaNet and SqueezeNet from float32 to bfloat16 with real data slightly increases the data infeed percentage as well for similar reasons.
It also shows that performance can be improved when infeed time increases.

DenseNet and MobileNet have zero data infeed time.
Compared with ResNet, they train fewer images/second, putting less stress on the host to infeed data.
Switching from float32 to bfloat16 increases the performance of both workloads using real data.
Thus they are likely bottlenecked by memory bandwidth in the device.

Unlike CNNs, Transformer processes sequences, which are smaller than images and demand minimal computation for data decoding and/or preprocessing.
So Transformer does not have significant infeed time, as expected.
Unfortunately, its tensor2tensor implementation does not support synthetic data, so we omit the shaded bars for Transformer in \Fig{fig:casestudy}.

\paragraph{Architectural Implications}
Scaling performance of the CPU host to match the TPU device is crucial for utilization of the accelerator's computation resource.
For workloads limited by data infeed from the host to the device, resolving the bottleneck can improve performance by at least 34\%.
Such workloads include RetinaNet, ResNet-50, and SqueezeNet using bfloat16.
Sequence models such as Transformer do not stress data infeed as much as CNNs.
By increasing FLOPS utilization, data quantization can turn a compute-bound workload into one that is infeed-starved.
With a powerful CPU host, further data quantization can yield greater performance gain, if it is valid.
8-bit training is an example~\cite{banner2018scalable}.

\subsection{TPU v3}
\label{sec:tpu:v3}

\begin{figure}[t]
    \centering
        \subfloat[TPU v3 over v2]{\includegraphics[width=0.33\columnwidth]{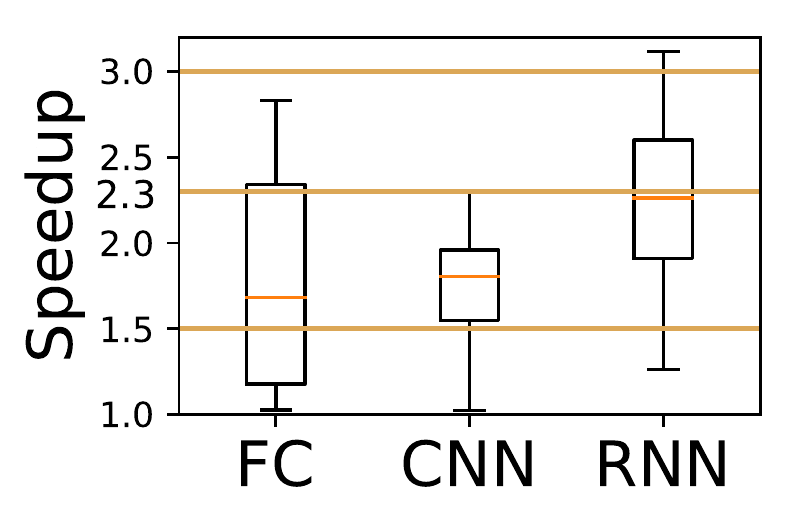}}
        \subfloat[FC Ops]{\includegraphics[width=0.33\columnwidth]{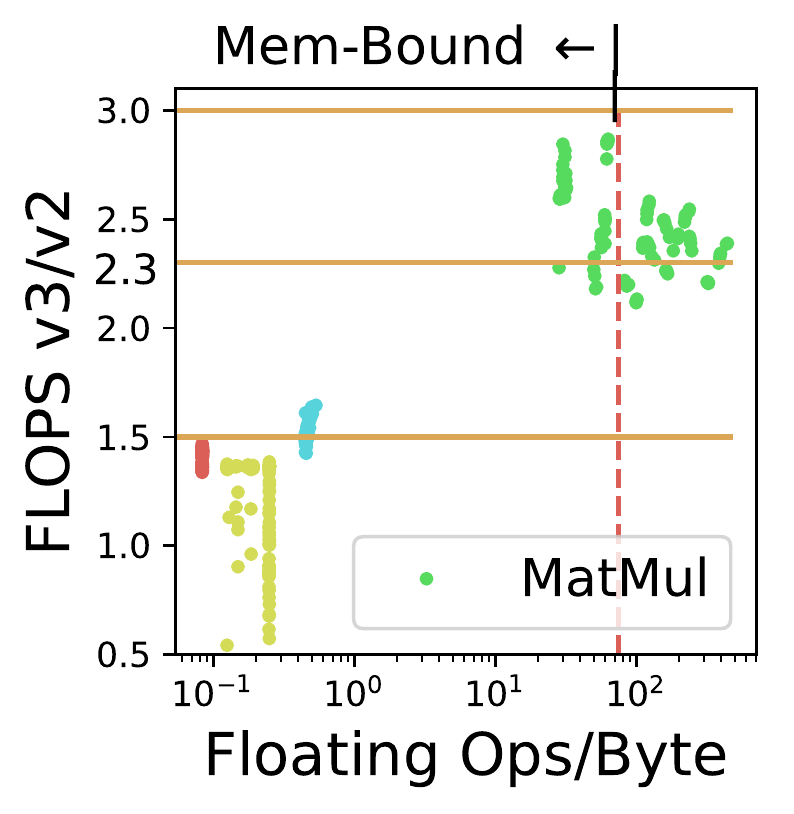}}
       \subfloat[CNN Ops]{\includegraphics[width=0.33\columnwidth]{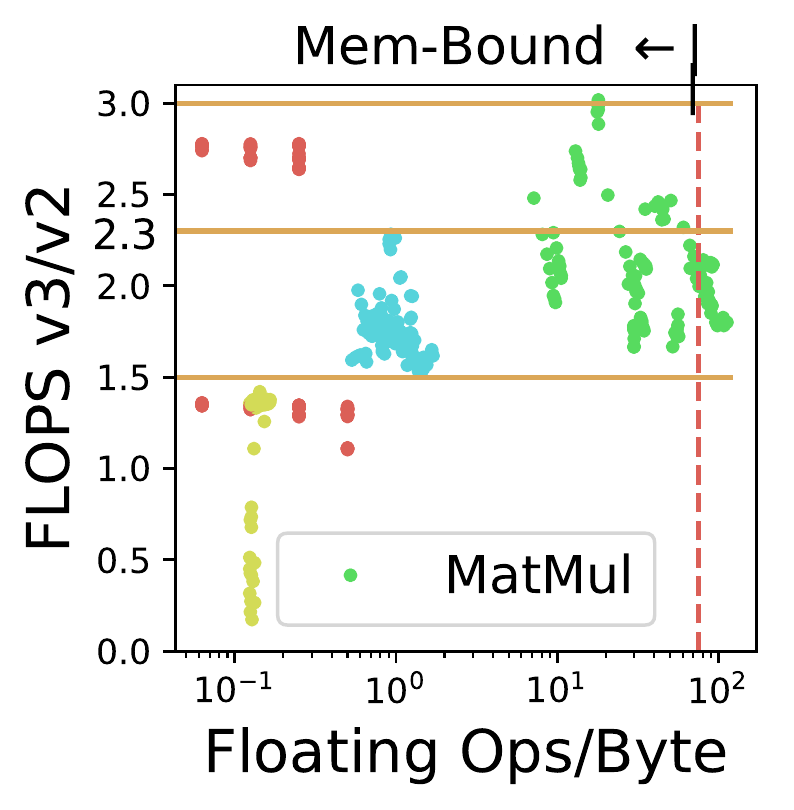}}
        \vspace{-0.5em}
    \caption{
    (a) Speedup of TPU v3 over v2 running end-to-end models.
    (b) and (c) Speedup comparison for FC and CNN operations.
    TPU v3's larger memory supports doubled batch sizes,
    so memory-bound operations have triple speedup if they benefit from larger batch size, and 1.5$\times$ speedup if not. 
    Compute-bound on v3 operations have 2.3$\times$ the speedup.
    The red line (\SI{75}{Ops/Byte}) is the inflection point in the TPU v2 roofline. (See roofline and legends in Fig~\ref{fig:tpu_rooflines}.)}
    \vspace{-1.5em}
    \label{fig:tpuv3}
\end{figure}

This section focuses on the differences between TPU v2 and v3.
\Fig{fig:tpuv3} compares TPU v3 and v2 using FC, CNN with bottleneck block, and basic RNN models. Batch size for v3 is twice that for v2, thanks to its doubled memory capacity.
\Fig{fig:tpuv3}(a) shows the speedups of end-to-end ParaDnn models.
Because end-to-end model speedup depends on operations, we first discuss the operation breakdown in detail.
\Fig{fig:tpuv3}(b)--(c) show arithmetic intensity on the $x$-axis and the speedup of FC and CNN operations on the $y$-axis.
Data points are colored by operation types, consistently with \Fig{fig:tpu_rooflines}(b) and (d).
As a reference, the red dashed line is the inflection point in the TPU v2 roofline from~\Fig{fig:tpu_rooflines}, where arithmetic intensity is \SI{75}{Ops/Byte} (\SI{180}{TFLOPS} / \SI{2.4}{TB/s}).
The operations on the left of the red line are memory-bound, and the ones on the right are compute-bound.
We can group the operations in four classes, as follows.

\paragraph{Compute-Bound Ops}
The peak FLOPS of TPU v3 is 2.3$\times$ that of v2, so compute-bound operations are improved by about 2.3$\times$ on v3.
Such operations are on the right of the red dashed line in \Fig{fig:tpuv3}(b).

\paragraph{Memory-Bound Ops (2$\times$ batch size)}
The maximum speedup of the memory-bound operations (mainly the MatMuls in \Fig{fig:tpuv3}(b)--(c)) is 3$\times$. 
The tripled speedup comes from doubled batch size (enabled by doubled memory capacity) and memory bandwidth improvement.
Thus we can infer v3 has 1.5$\times$ bandwidth improvement (\SI{3.6}{TB/s} per board) over v2, although its memory bandwidth has not been officially disclosed.
This is because on the slanted line of a roofline model, doubled batch size means doubled arithmetic intensity, and thus doubled FLOPS, because the ratio of FLOPS to arithmetic intensity is fixed.
And switching from v2's roofline to v3's gives a FLOPS improvement equal to the bandwidth improvement.
The fact that the overall speedup is 3$\times$ indicates that the bandwidth improvement is $3/2 = 1.5\times$.

\paragraph{Other Memory-Bound Ops}
The 1.5$\times$ bandwidth improvement assumption is corroborated by the 1.5$\times$ speedup of other memory-bound operations, represented by the non-MatMul FC operations in the lower left corner of \Fig{fig:tpuv3}(b).
The performance of those operations does not increase with larger batch size, as shown by the vertical alignment of each operation type in \Fig{fig:tpu_rooflines}(b).
Thus the 1.5$\times$ performance improvement in \Fig{fig:tpuv3}(b) is from bandwidth improvement.

\paragraph{Boundary Cases}
The compute-bound MatMuls in \Fig{fig:tpuv3}(c) become memory-bound on TPU v3, so the speedup is $< 2.3\times$.
Such operations have arithmetic intensity between 75 and 117,
because the roofline inflection point of v3 is at $x = 420 / (2.4*1.5) = 117$.
CrossReplicaSum (yellow dots) is slowed down on TPU v3, which may be because of more replicas across more MXUs.

\paragraph{End-to-End Models}
In \Fig{fig:tpuv3}(a) the maximum speedups are 2.83$\times$ (FC), 2.31$\times$(CNN), and 3.11$\times$(RNN).
Speedup increases with model width (second column of \Tbl{table:var_range}), and the maximum speedup is achieved by the largest width.
FCs with close to 3$\times$ speedup are dominated by memory-bound MatMuls.
Exceptions are RNNs with more than 3$\times$,; these have the largest embedding size (900), indicating that TPU v3 optimizes large embedding computations.


\paragraph{Architectural Implications}
ParaDnn enables users to exam a wide range of workloads, from memory-bound to compute-bound.
Compared to v2, TPU v3 shows three main levels of speedup: 2.3$\times$ for compute-bound operations, 3$\times$ for memory-bound MatMuls, and 1.5$\times$ for other memory-bound operations.
This is the result of its 2.3$\times$ FLOPS, 2$\times$ memory capacity, and 1.5$\times$ memory bandwidth.
For architects, the relative improvement of FLOPS and memory is a trade-off based on key workloads and budgets.


\section{Cross-Platform Comparison}
\label{sec:xcompare}

\begin{figure}[t]
    \centering
        \subfloat[CPU]{\includegraphics[width=0.3\columnwidth]{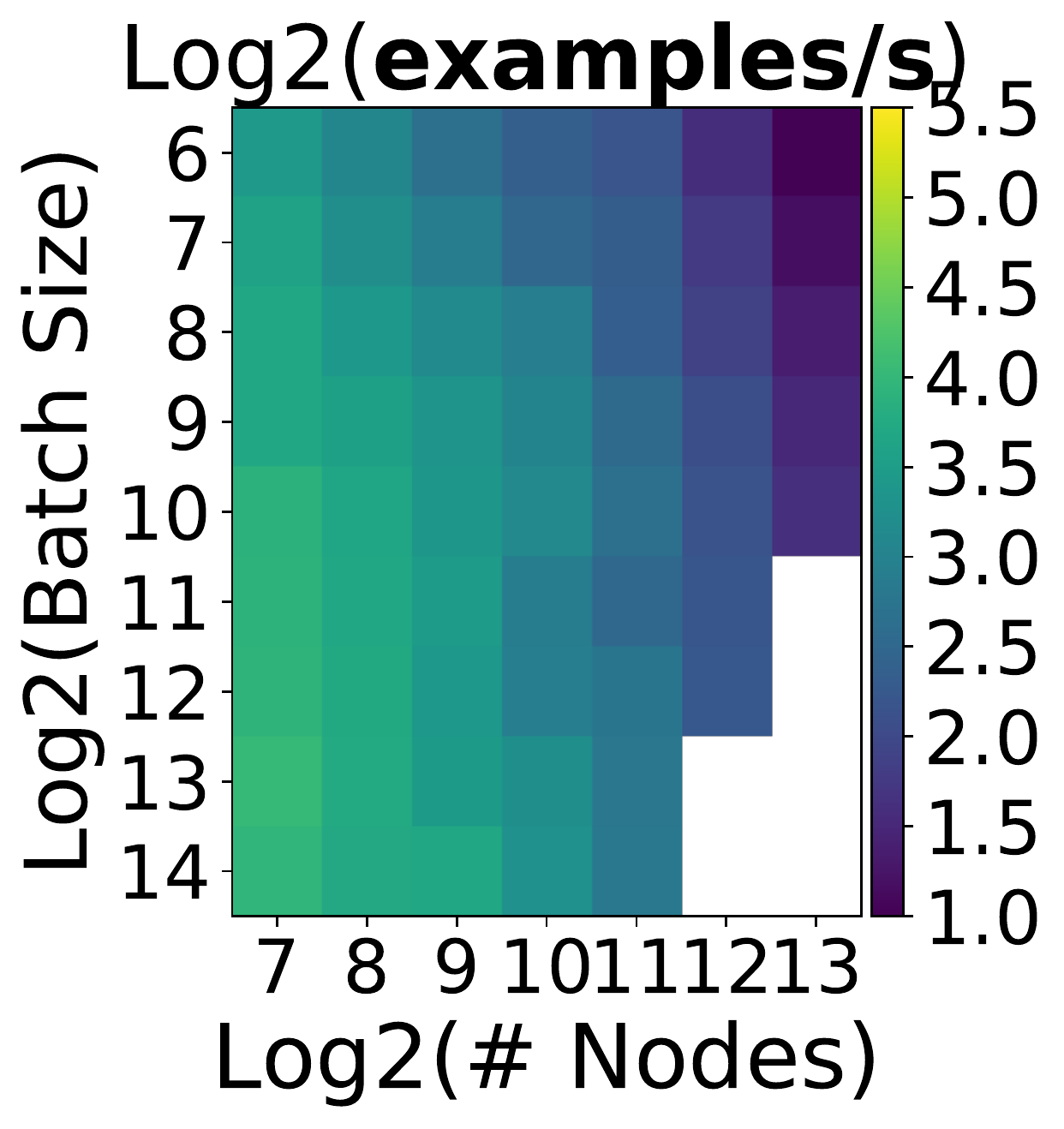}}
        \subfloat[GPU]{\includegraphics[width=0.3\columnwidth]{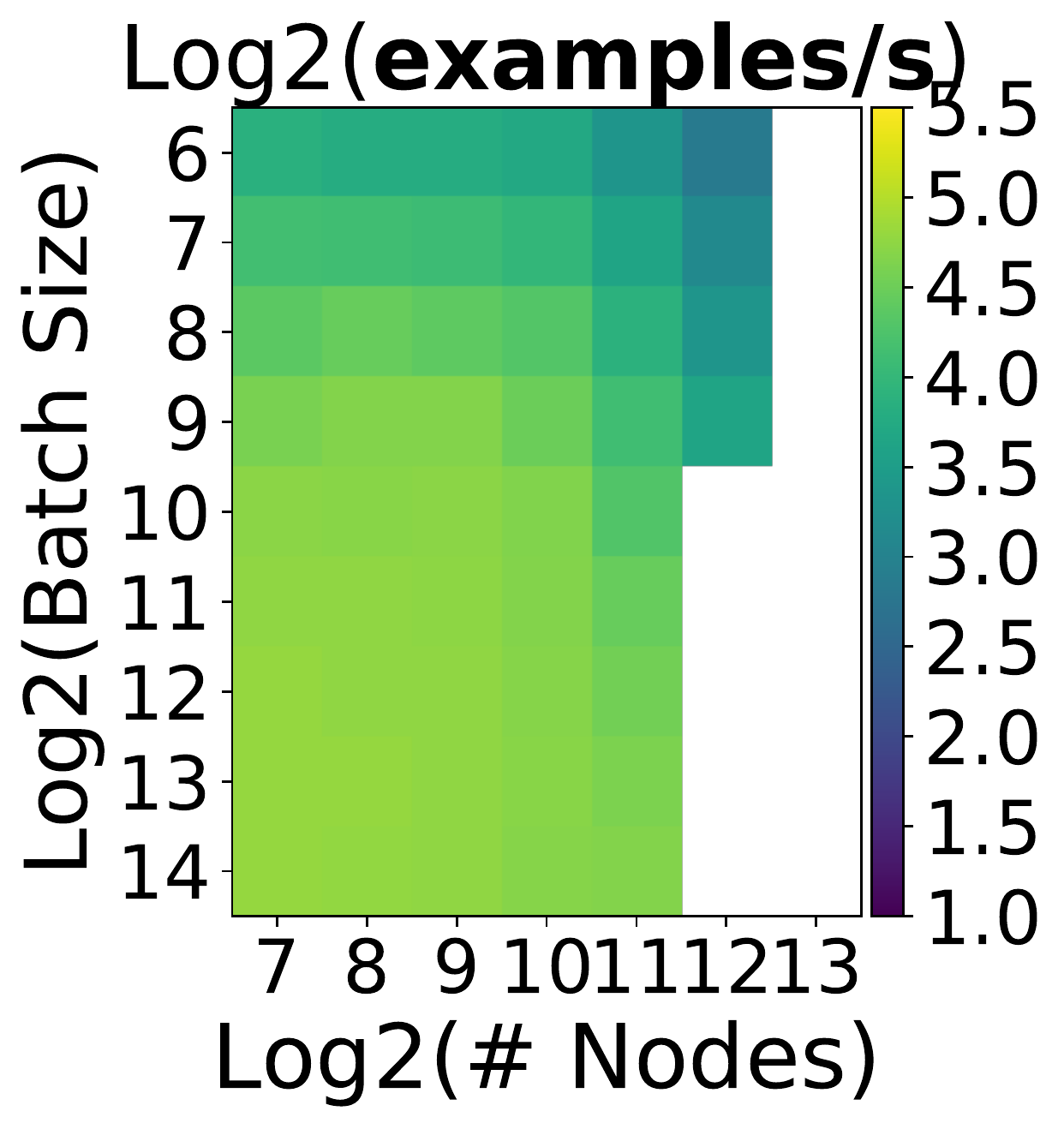}}
        \subfloat[TPU]{\includegraphics[width=0.3\columnwidth]{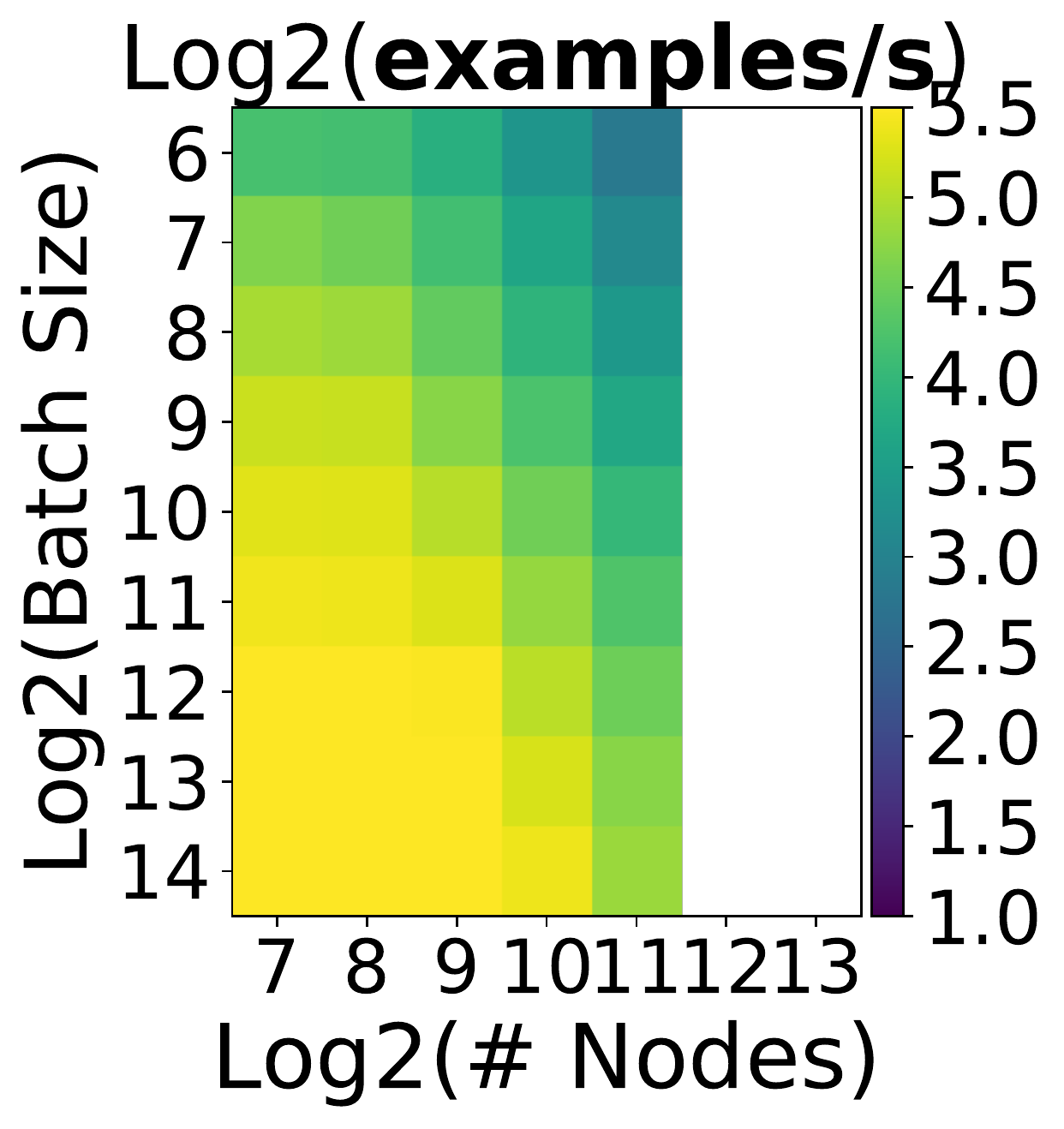}}
        \vspace{-0.5em}
    \caption{Examples/second of fully-connected models with fixed layer (64). Examples/second decreases with nodes and increases with batch size. White squares indicate models that encounter out-of-memory issues. The CPU platform runs the largest model because of its large memory.}
    \vspace{-1em}
    \label{fig:fc_example}
\end{figure}

In this section, we conduct cross-platform comparison using TPU, GPU, and CPU, so that users can choose the most suitable platform based on models of interest.
We find that there are scenarios where each of the platforms is valuable, trading off flexibility and specialization.
We also discuss the implications for future architecture designs.
The following is a summary of the key takeaways:

\begin{itemize}
\vspace{-0.5em}
\item \textbf{TPU}
is highly-optimized for large batches and CNNs, and has the highest training throughput.
\vspace{-0.5em}
\item \textbf{GPU}
shows better flexibility and programmability for irregular computations, such as small batches and non-MatMul computations.
The training of large FC models also benefits from its sophisticated memory system and higher bandwidth.
\vspace{-0.5em}
\item \textbf{CPU}
has the best programmability, so it achieves the highest FLOPS utilization for RNNs, and it supports the largest model because of large memory capacity.
\end{itemize}
\vspace{-0.5em}

We consider two performance metrics, examples/second and speedup.
Examples/second measures the number of examples trained per second, which is throughput.
We use it as a proxy for end-to-end performance.
The speedup of one platform over another is the ratio of the former's performance (examples/second) over the latter's.

\subsection{Fully-Connected DNNs}
\label{sec:xcompare:fc}

This subsection provides systematic analysis of the performance and speedups for fully-connected (FC) models.


\begin{figure}[t]
    \centering
        \subfloat[LR Weights]{\includegraphics[width=0.4\columnwidth]{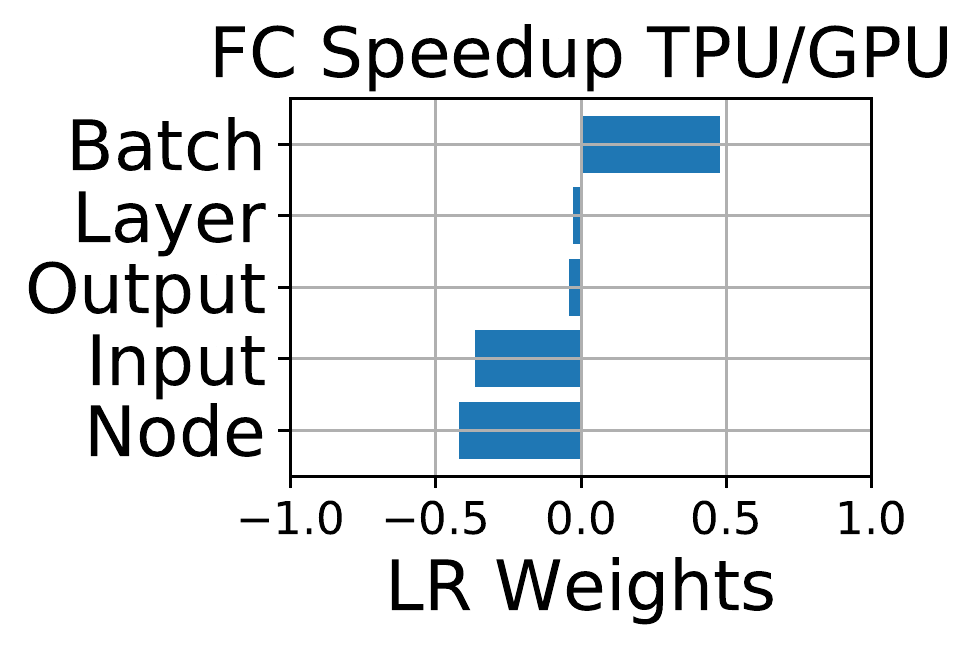}}
        \subfloat[Batch Size]{\includegraphics[width=0.3\columnwidth]{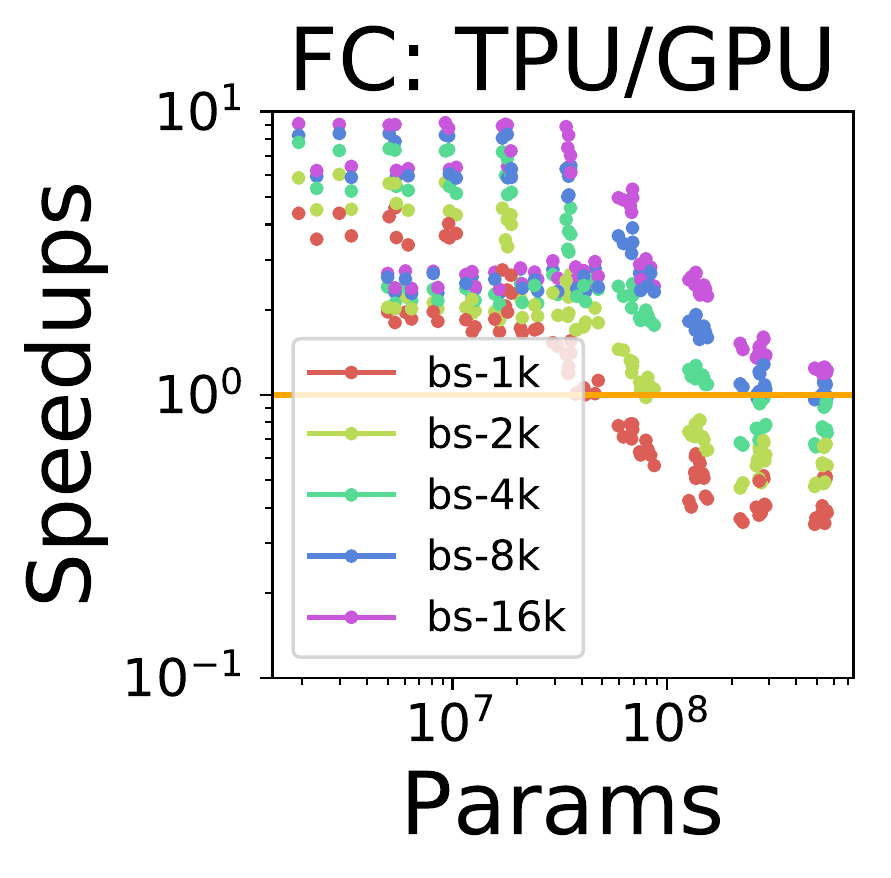}}
        \subfloat[Node]{\includegraphics[width=0.3\columnwidth]{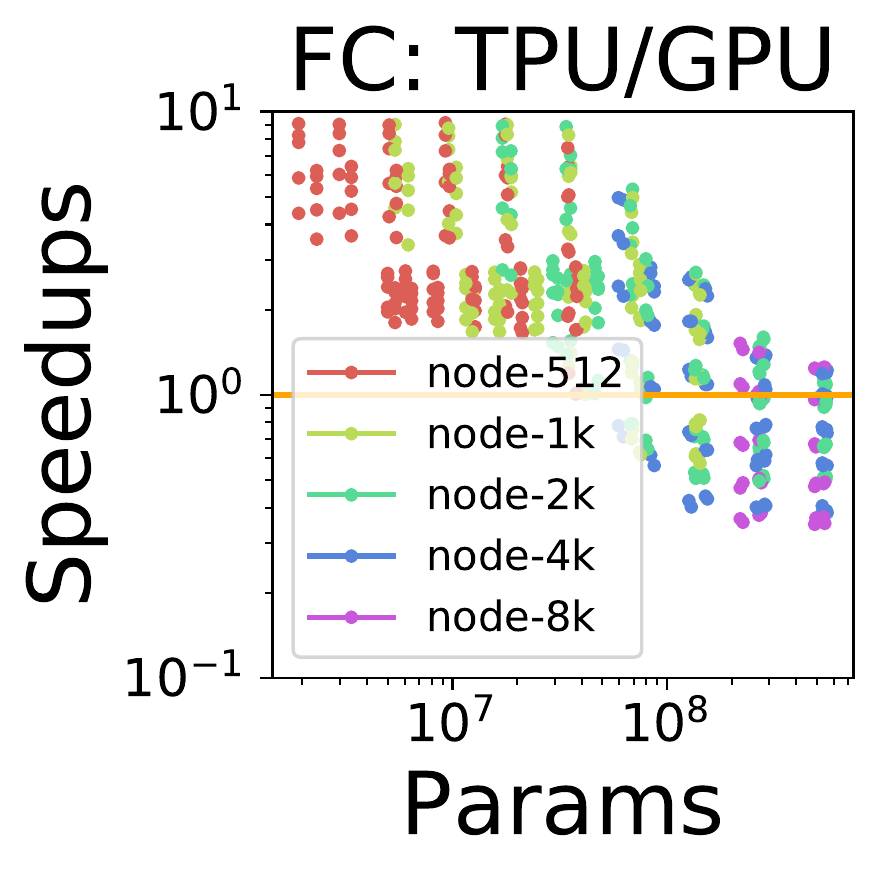}}
        \vspace{-0.5em}
     \caption{Small FC models with large batch sizes prefer TPU, and large models with small batch sizes prefer GPU, indicating systolic arrays are better with large matrices, and the warp scheduling on GPU is more flexible for small matrices.}
     \vspace{-1.5em}
    \label{fig:fc_speedup_tpu_gpu}
\end{figure}

\paragraph{Examples/second}
Figure~\ref{fig:fc_example} shows throughput for varying node counts and batch sizes but fixed layer count (64).
We use LR weights introduced in \Sec{sec:tpu:flops} to quantify the hyperparameter effects (not shown owing to space limitations).
Layer and node counts have negative weights, because it is time consuming to train large models with many layers and nodes.
Batch size greatly improves examples/second on GPU and TPU, but not CPU, because 
the parallelism available with small batch sizes is enough to highly utilize CPU.

It is interesting to note that only the CPU supports the largest models, and the GPU supports larger models than the TPU.
This is because every hardware core keeps one copy of the model, so the largest model supported is determined by memory per core, as explained in \Sec{sec:platform}.
In Figure~\ref{fig:fc_example}, the white squares indicate models that encounter out-of-memory (OOM) issues.
CPU has the highest memory per core (120 GB), and GPU (16 GB) is higher than TPU (8~GB).
While TPUs and GPUs may draw more attention, as of today the only choice for extremely large models is the CPU, which supports all model sizes.
For example, Facebook reports using dual-socket, high-memory CPU servers to train ranking models
for News Feed and to perform anomaly detection (Sigma), both of which are 
fully-connected networks~\cite{hazelwood2018applied}.
That fact emphasizes the need for model parallelism and pipelining~\cite{dean2012large,jia2018beyond,gpipe} on GPU and TPU, such that those powerful accelerators can support larger models.


\paragraph{TPU over GPU Speedup}
To further investigate the best hardware platform for an FC model, we analyze TPU over GPU speedups.
Figure~\ref{fig:fc_speedup_tpu_gpu}(a) plots the linear regression weights across FC hyperparameters for TPU over GPU speedup.
To show the design space of FC models, 
Figures~\ref{fig:fc_speedup_tpu_gpu}(b)--\ref{fig:fc_speedup_tpu_gpu}(c) are
scatter plots showing numbers of model parameters on the $x$~axis and speedups on the $y$~axis.
To display the effects of the hyperparameters, we color code data points to reflect batch size (Figure~\ref{fig:fc_speedup_tpu_gpu}(b)) and node count (Figure~\ref{fig:fc_speedup_tpu_gpu}(c)).
Overall, 62\% of the FC models perform better on TPU (speedup $> 1$).

TPU is well suited for large batch training, because systolic arrays are very good at increasing throughput~\cite{kung1982systolic}.
The positive weight in Figure~\ref{fig:fc_speedup_tpu_gpu}(a) and the horizontal color bands in Figure~\ref{fig:fc_speedup_tpu_gpu}(b)
show that large batch size is the key to higher TPU over GPU speedup.
This suggests that the matrix multiply units (MXU) of TPU, implemented with systolic arrays~\cite{jouppi2017datacenter,dean2017hotchips}, need large batches to reach full utilization.  
But GPU is a better choice for small batch sizes, because it executes computation in warps, so
it packs small batches and schedules them on stream multiprocessors more easily~\cite{nickolls2010gpu}.

GPU is a better choice for large models and datasets, 
suggesting that it is more optimized for large FC memory reuse/streaming requirements.
Large models and datasets lower speedups, shown by the negative weights of node count, layer count, and input in Figure~\ref{fig:fc_speedup_tpu_gpu}(a) and the scatter plot Figure~\ref{fig:fc_speedup_tpu_gpu}(c),
corroborated by the overall negatively-correlated trend of speedup with number of parameters in Figure~\ref{fig:fc_speedup_tpu_gpu}.
FC models have minimal weight reuse and
large models have more weights,
so they put a lot of pressure on the memory system.
GPU has a more mature memory system and higher memory bandwidth than TPU, which makes GPU better-suited for the memory requirements of large FC models.

\begin{figure}[t]
    \centering
        \subfloat[LR Weights]{\includegraphics[width=0.4\columnwidth]{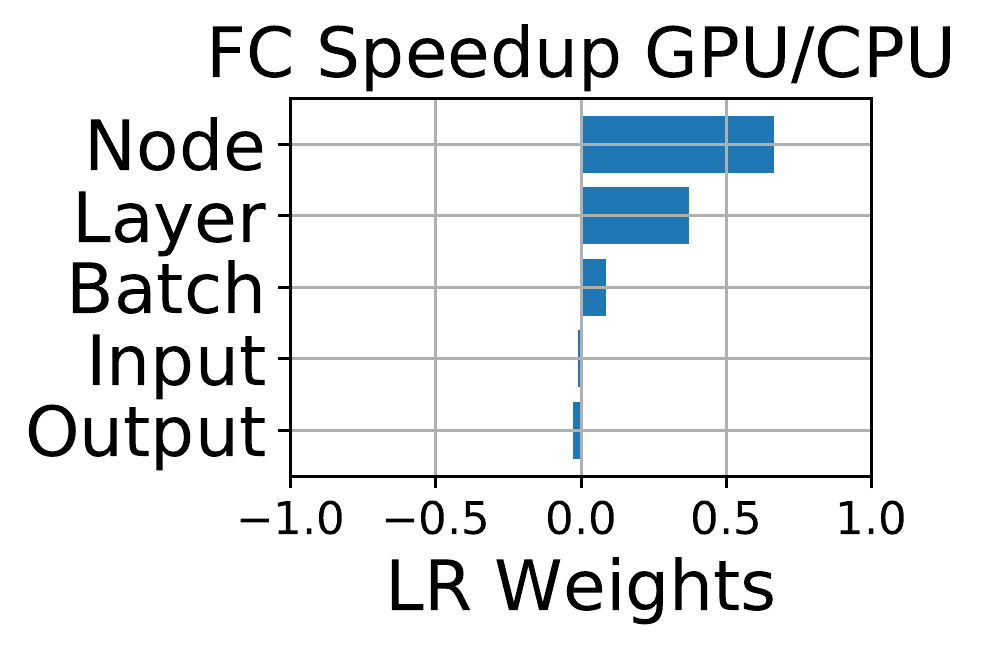}}
        \subfloat[Node]{\includegraphics[width=0.3\columnwidth]{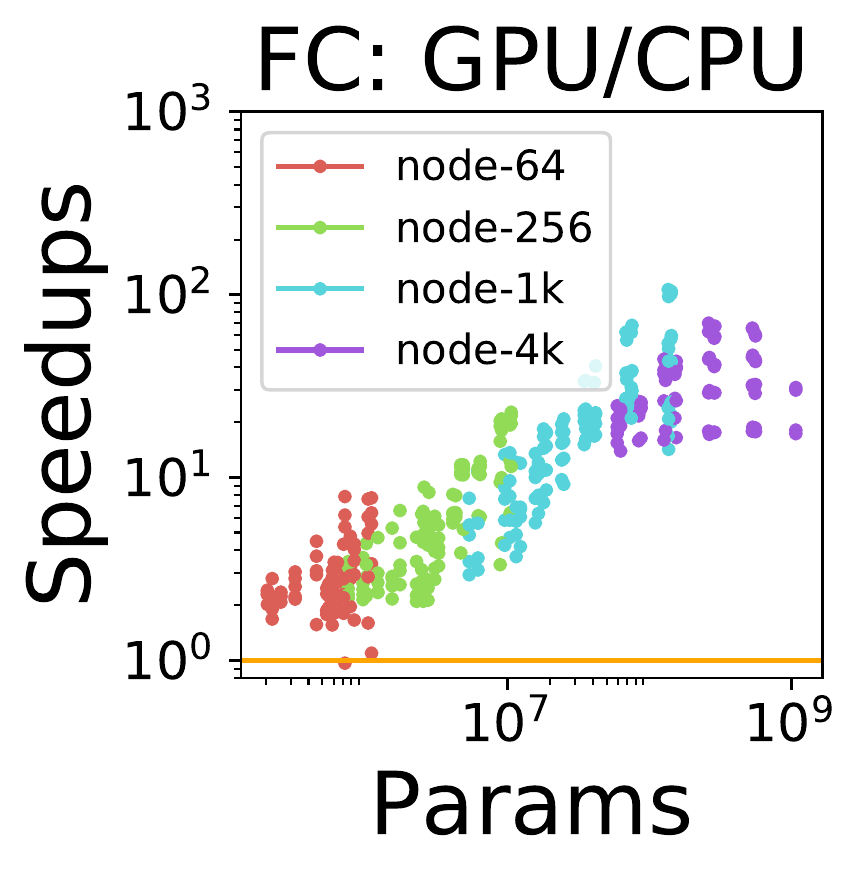}}
        \vspace{-0.5em}
     \caption{Large FC models with large batch sizes are
    better suited for GPU than CPU because the GPU's architecture can better utilize the extra parallelism.}
    \vspace{-1.5em}
     \label{fig:fc_speedup_gpu_cpu}
\end{figure}

\paragraph{GPU over CPU Speedup}
The speedup of GPU over CPU is an interesting comparison to TPU over GPU.
Figure~\ref{fig:fc_speedup_gpu_cpu}(a) shows the LR weights from learning GPU-over-CPU speedup. Figure~\ref{fig:fc_speedup_gpu_cpu}(b) shows the design space colored by node count .

GPU is a better platform for large FC models, because
its architecture is better at exploiting the extra parallelism from large batches and models.
As shown by Figure~\ref{fig:fc_speedup_gpu_cpu},
large models have higher speedups on GPU.
We also observe that large FC models prefer GPU over TPU,
witnessed by the positive trend in Figure~\ref{fig:fc_speedup_gpu_cpu}(b) and the negative trend in Figures~\ref{fig:fc_speedup_tpu_gpu}(b)--\ref{fig:fc_speedup_tpu_gpu}(c).
So GPU is the best platform for large FC models, but models with large batch sizes perform best on TPU, and better on GPU than on CPU.

\subsection{CNN and RNN}
\label{sec:xcompare:cnnrnn}

We now describe the speedup of CNNs and RNNs. 
Since our conclusions for CPUs and the hyperparameter LR weights on examples/second are similar to those in the previous section, we omit those results in the interest of brevity.



\begin{figure}[t]
    \centering
        \subfloat[LR Weights]{\includegraphics[width=0.4\columnwidth]{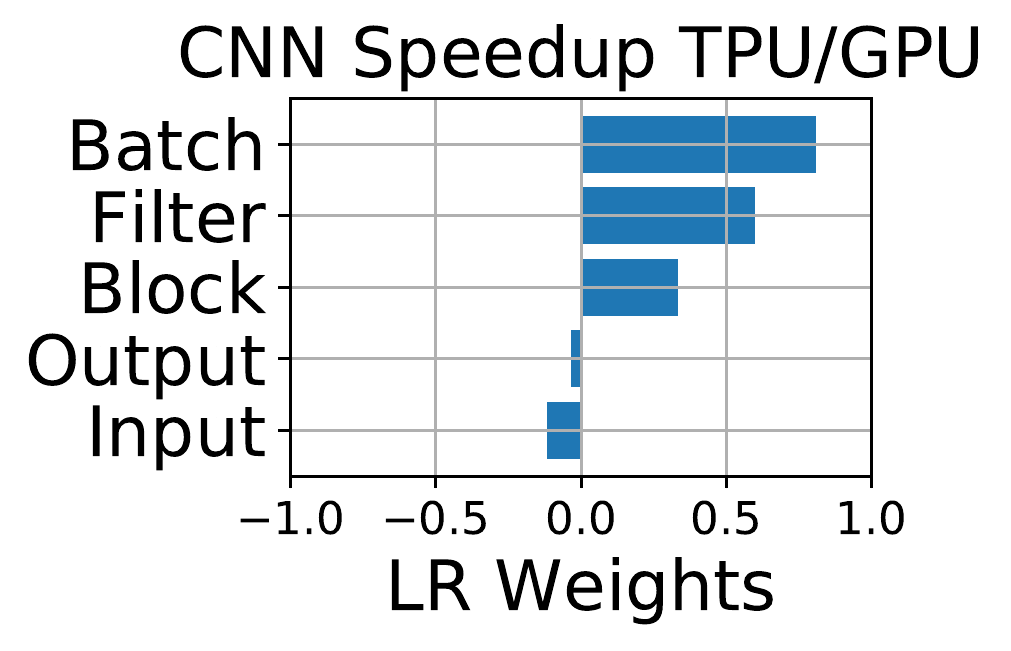}}
        \subfloat[Batch Size]{\includegraphics[width=0.3\columnwidth]{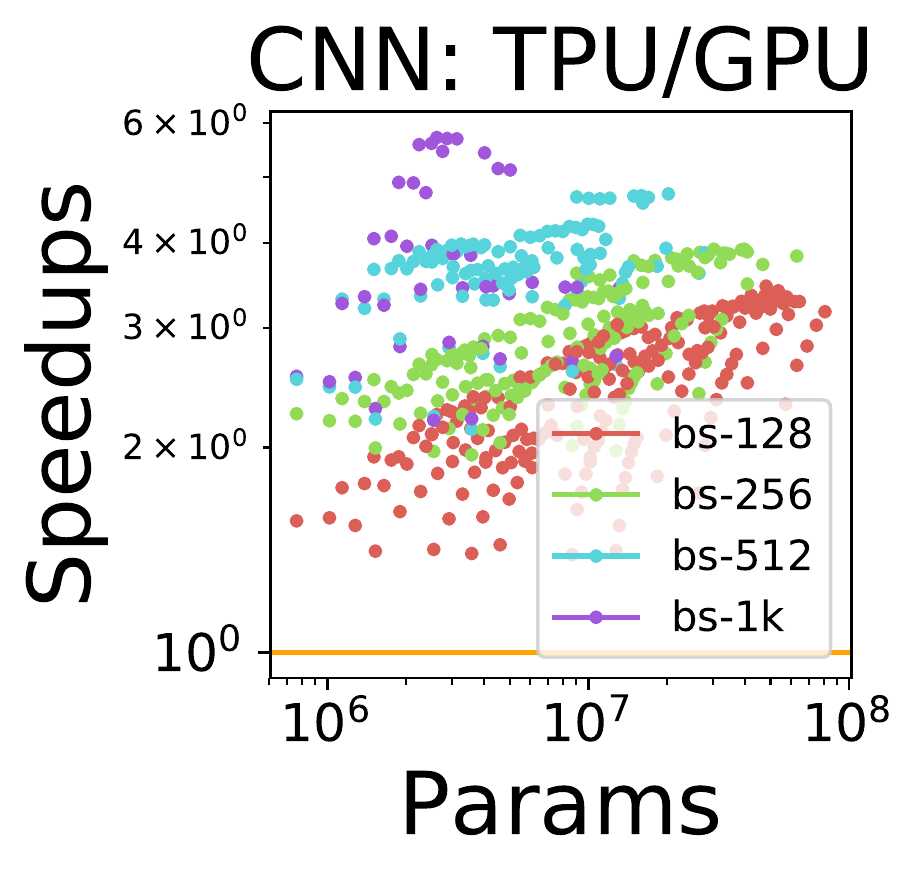}}
        \subfloat[Filter Size]{\includegraphics[width=0.3\columnwidth]{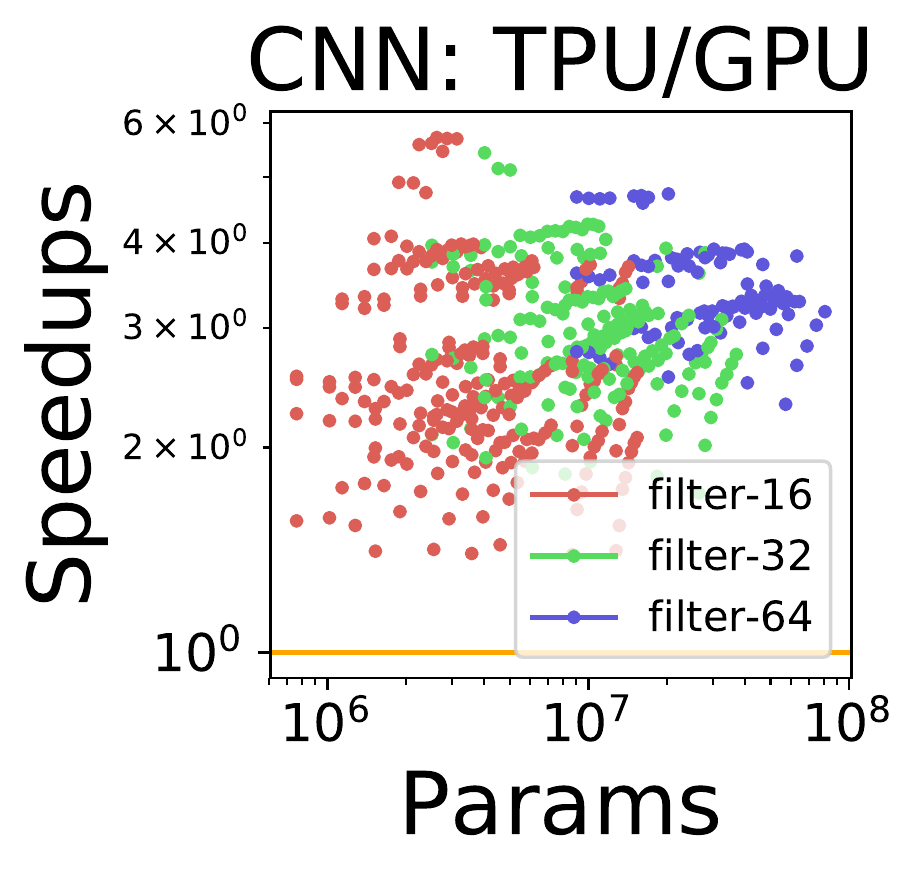}}
        \vspace{-1em}
        \qquad
        \subfloat[LR Weights]{\includegraphics[width=0.5\columnwidth]{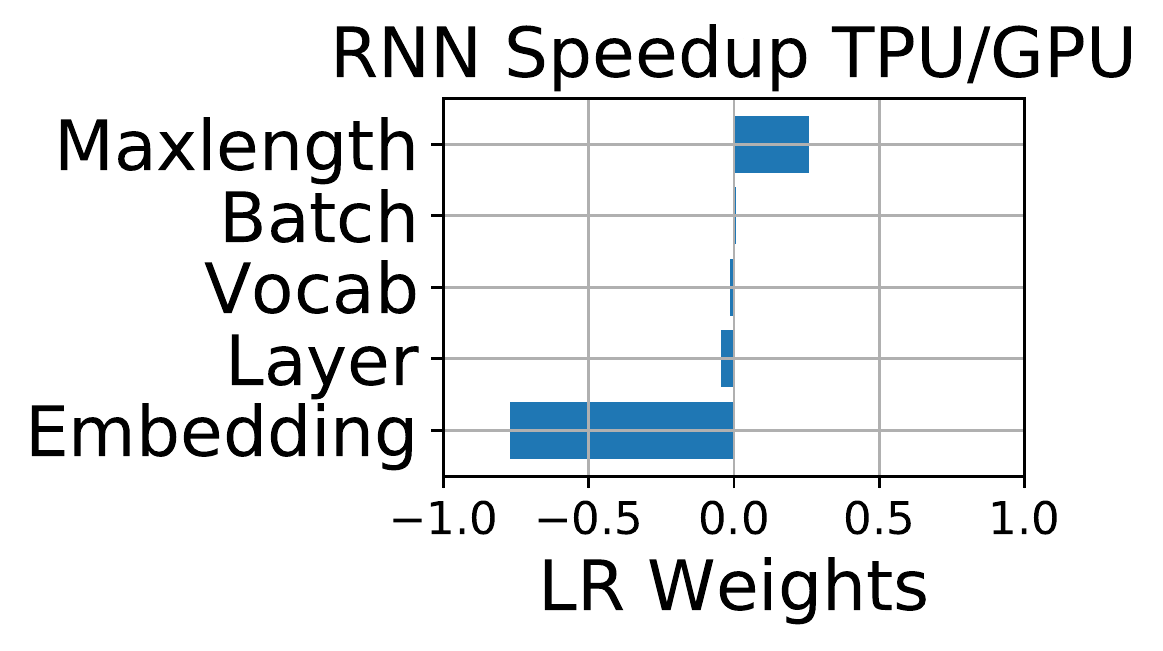}}
        \subfloat[Embedding Size]{\includegraphics[width=0.3\columnwidth]{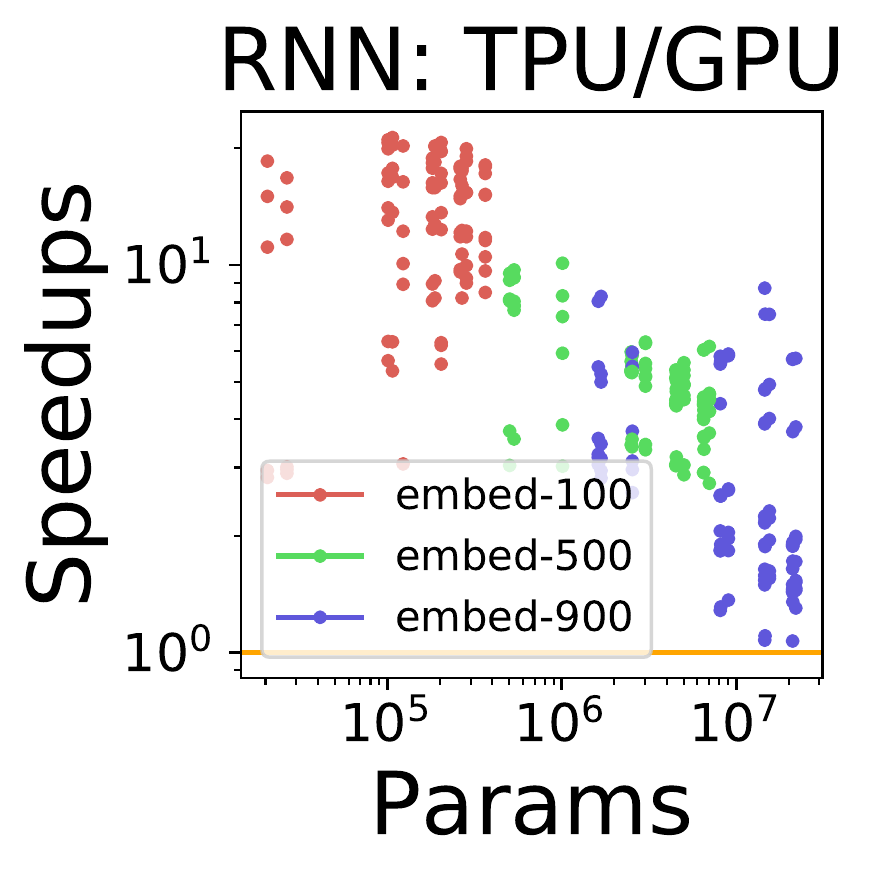}}
        \vspace{-0.5em}
    \caption{(a)--(c) TPU is a better choice than GPU for large CNNs, suggesting that TPU is highly-optimized for CNNs. (d)--(e) While TPU is a better choice for RNNs, it is not as flexible as GPU for embedding computations.}
    \vspace{-1em}
    \label{fig:cnnrnn_speedup_tpu_gpu}
\end{figure}

\paragraph{CNN}
Figures~\ref{fig:cnnrnn_speedup_tpu_gpu}(a)--\ref{fig:cnnrnn_speedup_tpu_gpu}(c) show the speedups of TPU over GPU.
All CNNs perform better on TPU.
Batch size is still the key to better TPU over GPU speedup for CNNs, shown by its positive LR weight in \Fig{fig:cnnrnn_speedup_tpu_gpu}(a) and the increasing speedup with batch size in \Fig{fig:cnnrnn_speedup_tpu_gpu}(b).

TPU is the best platform for large CNNs, suggesting that the TPU architecture is highly optimized for
the spatial reuse characteristics of
CNNs.
This is shown by the positive weights in Figures~\ref{fig:cnnrnn_speedup_tpu_gpu}(a) and \ref{fig:cnnrnn_speedup_tpu_gpu}(c),
where models with more filters and blocks have higher speedups.
It is different from \Sec{sec:xcompare:fc}, showing that TPU is not preferred for large FCs.
This suggests it is easier for TPU to optimize for large CNNs than large FCs, which may be because CNNs reuse weights.
FC models barely reuse weights, which introduces more memory traffic.
GPU is a feasible choice for small CNNs.
These conclusions only apply to single-GPU performance;
the multi-GPU case may be different.

\paragraph{RNN}
Figures~\ref{fig:cnnrnn_speedup_tpu_gpu}(d)--\ref{fig:cnnrnn_speedup_tpu_gpu}(e) show the speedup of TPU over GPU.
We display the embedding size in \Fig{fig:cnnrnn_speedup_tpu_gpu}(e), because the magnitude of its weight is greatest in \Fig{fig:cnnrnn_speedup_tpu_gpu}(d).
Embedding size has negative weights in \Fig{fig:cnnrnn_speedup_tpu_gpu}(d) and
embedding computation is more sparse than matrix multiplication.
This suggests that TPU is less flexible for doing non-MatMul computations than GPU.
TPU is better at dense computations like MatMuls.
Even so, RNNs are still up to 20$\times$ faster on TPU.
Optimizing non-MatMul computations is another opportunity for TPU enhancement.

\subsection{Overall Comparison}
\label{sec:xcompare:overall}

This section summarizes the speedup of TPU over GPU and the FLOPS utilization of all parameterized and real 
models. We do not show the results of using CPUs to train CNNs, because it is extremely time consuming and unlikely to 
contribute additional insights.

\begin{figure}[t]
    \centering
     \subfloat{\includegraphics[width=1\columnwidth]{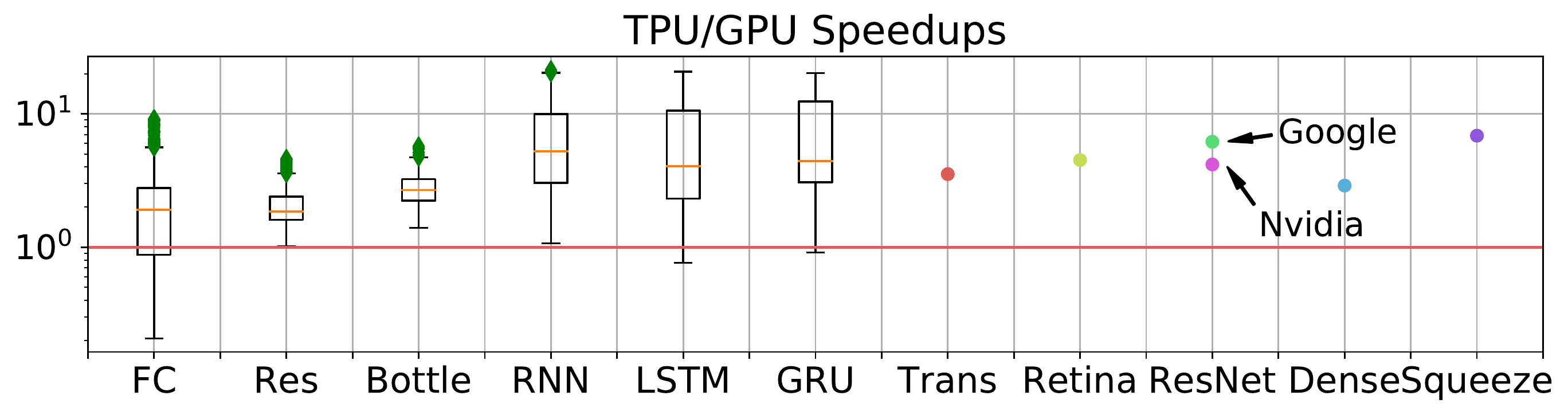}}
     \vspace{-1em}
     \qquad

  \subfloat{\includegraphics[width=1\columnwidth]{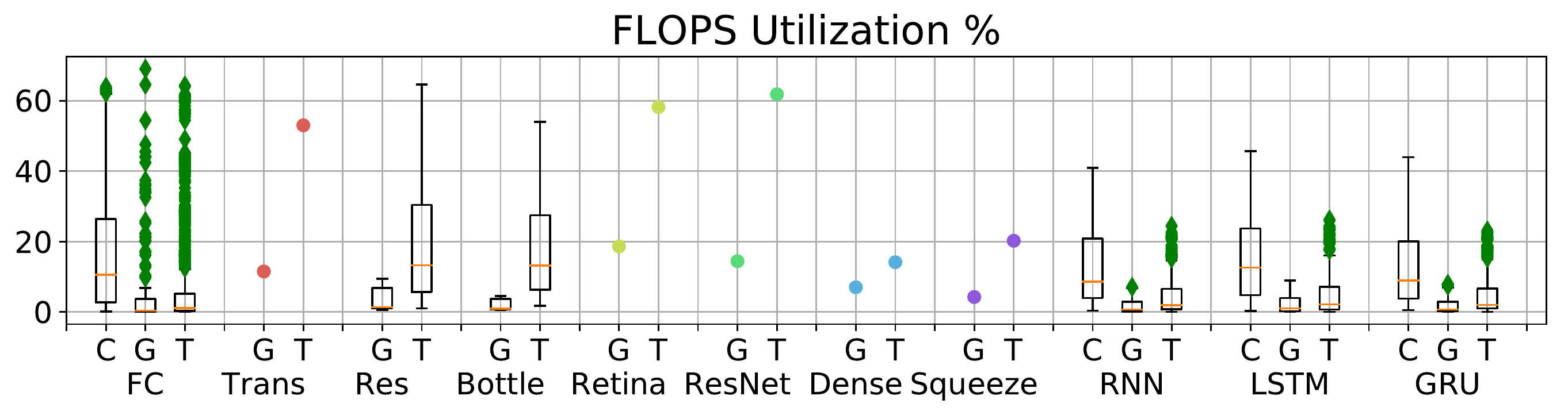}}
  \vspace{-1em}
    \caption{(Top) TPU over GPU speedups of all workloads. Note that the real workloads use larger batch sizes on TPU than on GPU. The NVIDIA version of ResNet-50 is from~\cite{case2018volta}. (Bottom) FLOPS utilization comparison for all platforms.}
    \vspace{-1em}
    \label{fig:overall}
\end{figure}

\paragraph{TPU over GPU Speedup}
\Fig{fig:overall}(top) summarizes the TPU over GPU speedups of all models.
Note that the real workloads use larger batch sizes on TPU than on GPU.
Speedup of TPU over GPU depends heavily on the nature of the workload measured.
The speedup of parameterized models has large ranges, from less than 1 to 10$\times$,
while the speedup of real workloads range from 3$\times$ (DenseNet) to 6.8$\times$ (SqueezeNet).
ParaDnn represents a more complete view of potential workloads, and
each real workload represents the concerns of certain users.
Benchmarking platforms with two kinds of workloads offer a more systematic
understanding of their behavior than those with only one kind.

To further compare TPU and GPU while relaxing the constraint on the software stack of the GPU, we also include the speedup relative to GPU performance of ResNet-50, reported in NVIDIA's Developer Blog~\cite{case2018volta} (annotated as NVIDIA in \Fig{fig:overall}(top)).
We note that NVIDIA's version of ResNet-50 uses unreleased libraries, and we were unable to reproduce the results. 
The speedup using ResNet-50 from Google is 6.2$\times$ compared to 4.2$\times$, which suggests software optimization can significantly impact performance.




\paragraph{FLOPS Utilization}
\Fig{fig:overall}(bottom) shows the FLOPS utilization of all workloads and platforms.
On average, the maximum FLOPS utilization of TPU is 2.2$\times$ that of GPU for all CNN models, and the ratio is 3$\times$ for RNNs.
The TPU FLOPS utilization of Transformers is consistent with FCs with 4k batch size, as shown in \Fig{fig:tpu_flops}.

For RNNs, TPU has less than 26\% FLOPS utilization and GPU has less than 9\%.
In contrast, CPU has up to 46\% utilization.
RNNs have irregular computations compared to FCs and CNNs,
due to the temporal dependency in the cells and the variable-length input sequences.
The parameterized RNNs are very basic, however.
Advanced RNN optimizations may be able to increase utilization on GPU and TPU.

ResNet-50 and RetinaNet have higher FLOPS utilization than DenseNet and SqueezeNet.
The real workloads are ranked by number of trainable parameters, shown in \Fig{fig:model_params}.
DenseNet has lower utilization because it has fewer filters than ResNet-50.
DenseNet's maximum number of filters is 24~\cite{huang2017densely}, and the minimum of ResNet-50 is 64~\cite{he2016deep}.
SqueezeNet is designed specifically to have fewer parameters with the use of 1x1 filters~\cite{iandola2016squeezenet}.
Therefore, parallel operations represent a smaller portion of the whole workload.
As a consequence of Amdahl's law, the small models are unable to utilize the parallelism available on GPU or TPU.

ResNet-50 has higher FLOPS utilization than CNNs with bottleneck blocks.
This is because the parameterized CNNs keep the number of blocks the same in each group, while ResNet-50 has more blocks in groups with more filters, and that increases FLOPS.

\section{Software Stack Advances}
\label{sec:compiler}

Custom hardware for deep learning opens opportunities for dramatic library, toolkit, and compiler optimizations.
We now describe how different versions of TensorFlow (TF) and CUDA affect performance.
We study data type quantization with software versions, because it depends on software support. As a reminder, for all results in the previous sections, we
use the latest versions of each software stack with 16-bit quantization support.
Software versions are summarized in the legends of \Fig{fig:compiler}.
ParaDnn can reveal software optimization focus (e.g., TF 1.9 optimizes small-batch CNNs); we omit these details
for brevity.

\subsection{TensorFlow Versions and TPU Performance}
\label{sec:compiler:tpu}
\begin{figure}[t]
    \centering
 \subfloat{\includegraphics[width=1\columnwidth]{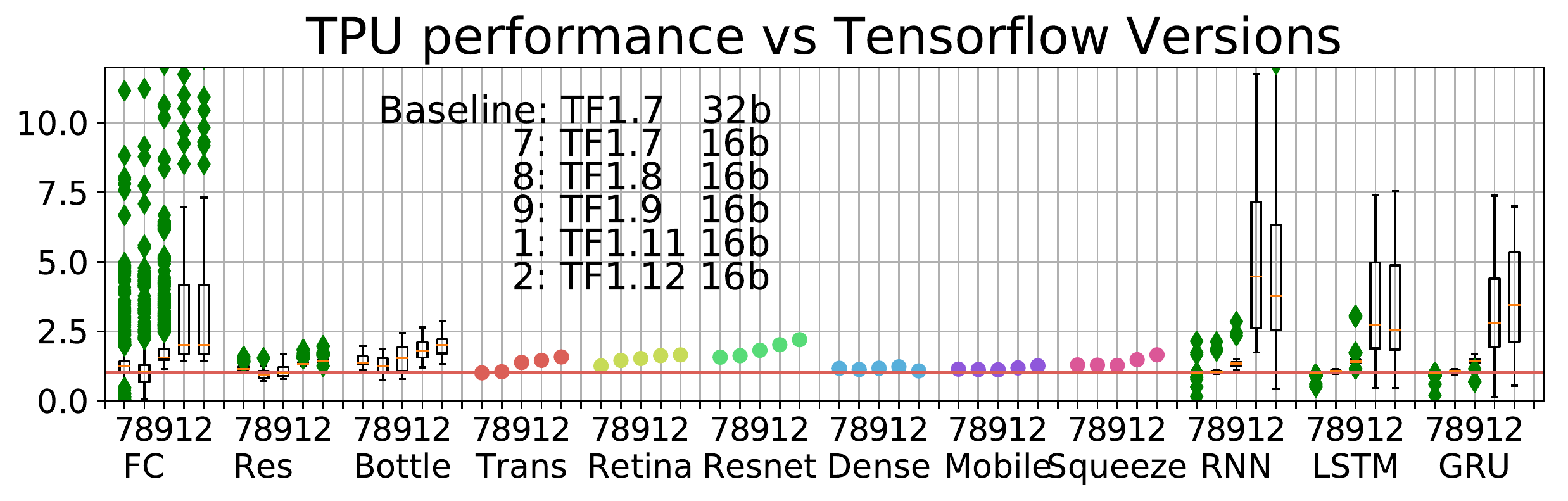}}
 \vspace{-1em}
    \qquad
    \subfloat{\includegraphics[width=0.73\columnwidth]{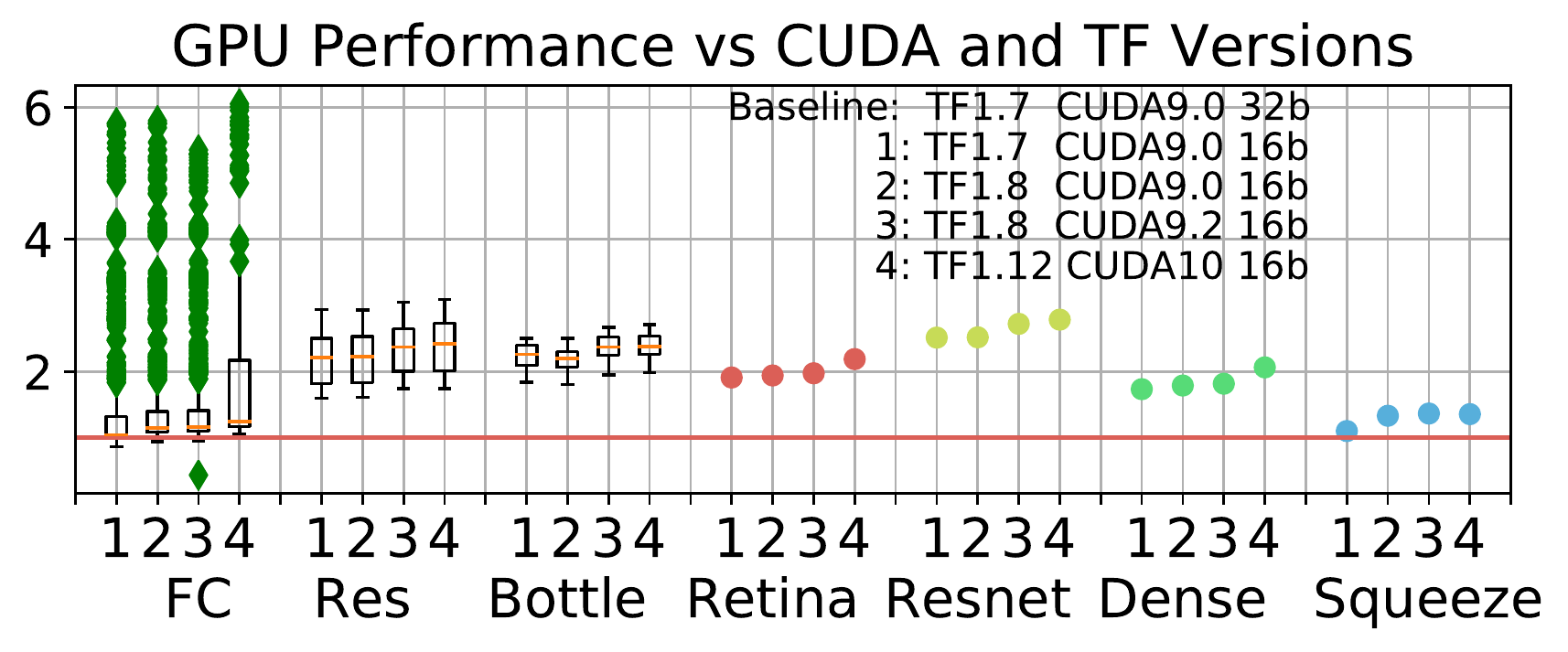}}
    \subfloat{\includegraphics[width=0.27\columnwidth]{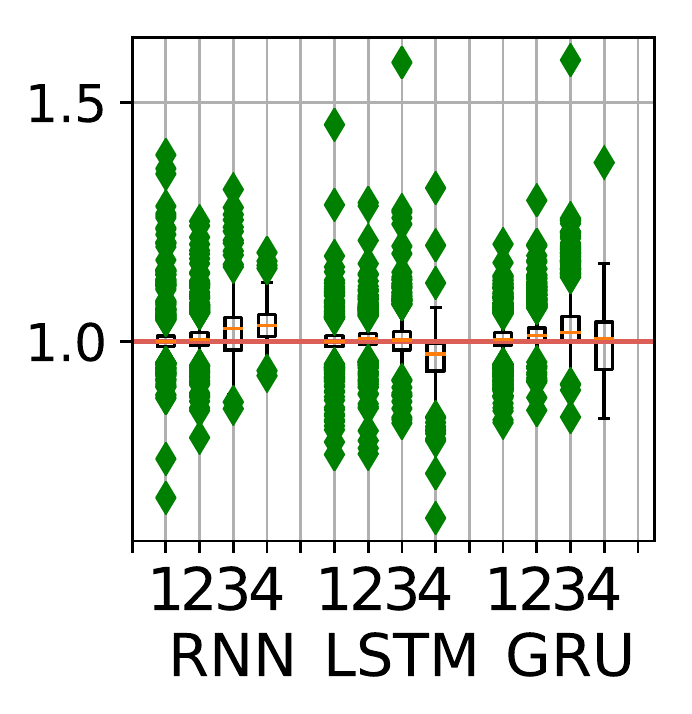}}
 \vspace{-1em}
    \caption{(a) TPU performance with TensorFlow updates. All ParaDnn models improve; Transformer, RetinaNet, and ResNet-50 improve steadily. (b) GPU speedups across versions of CUDA and TF. CUDA 9.2 improves CNNs more than other ParaDnn models, and ResNet-50 more than other real models. CUDA 10 does not improve RNNs or SqueezeNet.}
    \vspace{-1em}
    \label{fig:compiler}
\end{figure}

The compiler for the TPU is XLA~\cite{leary2017xla}, shipped with TF.
\Fig{fig:compiler}(a) shows TPU speedups obtained by running TF 1.7 to 1.12, treating 1.7 with float32 as the baseline.
The speedup is per model, maximizing batch size in each setting.
For example, using bfloat16 instead of float32 
allows larger batch size and thus higher speedup.%
\footnote{These experiments do not consider the impact of quantization on model accuracy.}
Moving from TF 1.7 to 1.12 improves performance for all ParaDnn models.
Although FC and CNN encounter performance regression with TF 1.8,
TF 1.9 fixes this anomaly and improves overall performance.

RNN performance is not improved much until TF 1.11.
TF 1.11 shows 10$\times$ speedup for RNN and 7.5$\times$ for LSTM and GRU.
Transformer, ResNet-50, and RetinaNet are improved continuously over TF updates.
Interestingly, SqueezeNet is improved starting from TF 1.11,
while the performance of DenseNet and MobileNet see little benefit.

In the 7 months (222 days) between the release of TF 1.7.0 (03/29/2018) and that of TF 1.12.0 (11/05/2018),
software stack performance improved significantly.
The 90th-percentile speedup of TPU is 7$\times$ for FC, 1.5$\times$ for Residual CNN, 2.5$\times$ for Bottleneck CNN, 9.7$\times$ for RNN, and 6.3$\times$ for LSTM and GRU.

The use of bfloat16 enables significant performance improvement for parameterized FC and CNN models.
90th-percentile speedups are up to 1.8$\times$ for FC and Bottleneck CNN, and 1.3$\times$ for Residual CNN.
Depending on the relative memory sizes of the data and model,
TPU can usually support doubled batch sizes by using 16 bits.
Transmitting 16 bits also relieves bandwidth pressure, which can speedup memory-bound operations as discussed in \Sec{sec:tpu:roofline} and \Sec{sec:tpu:systembalance}.
Larger performance increases may be possible with further reductions in bitwidth.

\subsection{CUDA Versions and GPU Performance}
\label{sec:compiler:gpu}

\Fig{fig:compiler}(b) shows GPU performance across versions of CUDA and TF.
The baseline is TF 1.7 and CUDA 9.0 with float32.
TF 1.8 does not improve GPU performance.
By lowering memory traffic and enabling larger batch sizes,  bitwidth reduction can speed up CNNs by more than 2$\times$.

We note that CUDA 9.2 speeds up ResNet-50 significantly more (8\%) than other real workloads (< 1\%).
CUDA 9.2 also speeds up ParaDnn CNNs more than FCs or RNNs.
CUDA 10 speeds up other models, 
but not SqueezeNet.
CUDA~10 also improves speedups for ParaDnn FCs and CNNs, but not as much for RNNs.
The overall 90th-percentile improvement for FCs is 5.2$\times$. For ParaDnn residual block and bottleneck block models it is 2.9$\times$ and 2.6$\times$, respectively.
In contrast, the 90-percentile improvement of parameterized models is 8.6\% for RNN, 3.5\% for LSTM, and 5.9\% for GRU.
The improvement from CUDA updates is less than that for TF updates on TPU, likely because CUDA and GPU platforms 
have matured greatly since becoming popular before 2010, while TPU v2 for training was only announced in May 2017.


\section{Limitations of this Work}
\label{sec:limitation}



\paragraph{Scope of this Work}
This work does not study DL inference, cloud overhead, multi-node systems, accuracy, or convergence. 
We intentionally leave these topics to future work, as each deserves in-depth study.
For example, evaluating inference entails different metrics, such as latency, and a different experimental setup, as network overhead may have a large effect.
\Sec{sec:tpu:systembalance} provides insight towards quantifying the network overhead, and 
we use synthetic data to minimize the cloud overhead, but virtualization, resource allocation, and job scheduling bring up more research questions.

NVIDIA's eight-node DGX-1 or Google's 256-TPU systems are not studied here.
Studying multi-node systems involves more system parameters, including numbers of nodes, inter-node bandwidth, inter-connect topology, and 
synchronization mechanisms. Cloud system overhead also becomes more acute in multi-node systems.

The validity of extrapolating training throughput to time-to-accuracy remains an open question. Recent work studied the number of training steps to accuracy as a function of batch sizes~\cite{shallue2018measuring}.
It shows that very large batch size results in sub-linear scaling, but the best batch size depends largely on the model and optimizer.
In a multi-node system, synchronization becomes more complicated, which results in different convergence behavior.

\paragraph{Tractability}
To keep the experiments tractable, we constrain the parameters in this work, including the ParaDnn hyperparameters (Table~\ref{table:var_range}) and the TPU iterations.
For example, we focus on large batches, as the platforms were designed for large batch training, and extremely small batches may lead to different conclusions.
We use the RMSProp optimizer, and SGD with momentum performs faster than RMSProp.

\section{Related Work}
\label{sec:relatedworks}

\paragraph{Benchmarks:}
``For better or worse, benchmarks shape a field,'' said David Patterson~\cite{patterson2012better}.
Indeed, benchmarks have been the driving force for compiler and architecture design for decades,
and notable examples include the SPEC CPU~\cite{henning2006spec} and PARSEC multiprocessor benchmarks~\cite{bienia2008parsec}.
Recently, work has focused on domain-specific benchmark suites including CortexSuite~\cite{thomas2014cortexsuite}, TonicSuite~\cite{hauswald2015djinn}, Sirius~\cite{hauswald2015sirius}, Fathom~\cite{adolf2016fathom}, DAWNBench~\cite{coleman2017dawnbench}, and MLPerf~\cite{mlperf}.
It is impossible to make any performance conclusions without benchmarks.

Benchmark designers must take care to avoid bias.
Existing benchmark suites come with limitations as discussed in \Sec{sec:synbench}.
ParaDnn is the first parameterized benchmark suite for deep learning in the literature.
In the same spirit as parameterized benchmarks, synthetic benchmarks have commonly been used, such as
BenchMaker~\cite{joshi2008return}, and SYMPO~\cite{ganesan2010system},
constructing benchmarks with hardware-independent characteristics.
Some try to match the statistical characteristics of real applications~\cite{aimatrix,kim2014workload}.
Synthetic approaches are common in domain-specific benchmarking, e.g., CAD~\cite{turki2012towards,stroobandt2000generating}, statistical network inference~\cite{schaffter2011genenetweaver}, and database~\cite{saleem2015feasible}.

\paragraph{Benchmarking}
Our use of deep learning models to compare up-to-date platforms, Google's TPU v2/v3 and NVIDIA's V100 GPU, distinguishes this work from previous cross-platform comparisons.
Shi et al. compare CPU (Intel i7-3820 and E5-2630v3) and GPU (GTX 980, GTX 1080, and K80) platforms and deep learning frameworks~\cite{shi2016benchmarking}.
Bahrampour et al. compare deep learning frameworks~\cite{bahrampour2016comparative}.
Others compare cloud computing providers~\cite{kothari2011comparison}, heterogeneous platforms~\cite{che2009rodinia}, and cloud support for HPC~\cite{he2010case}.

\section{Conclusion}
This paper provides a comprehensive benchmarking analysis of deep neural network training hardware and software, and valuable lessons learned for future system designs.
We present architectural bottlenecks of the TPU platform and provide suggestions for future improvement. 
Using ParaDnn, our parameterized benchmark suite for end-to-end deep learning, along with six
real-world models, we compare the hardware and software of the TPU, GPU, and CPU platforms.
We present several new observations and insights into the design of specialized hardware and software for
deep learning and motivate the need for further work in this field.




\section{Acknowledgement}
This work was supported in part by Google's TensorFlow Research Cloud (TFRC) program, NSF Grant
$\#$ CCF-1533737,
and the Center for Applications Driving Architectures (ADA), one of six centers of JUMP, a Semiconductor Research Corporation program co-sponsored by DARPA. 
The authors would like to thank
Frank Chen,
Blake Hechtman,
Jim Held,
Glenn Holloway, Dan Janni, Peter Mattson, Lifeng Nai,
David Patterson, Francesco Pontiggia, Parthasarathy Ranganathan,
Vijay Reddi,
Bjarke Roune,
Brennan Saeta, Zak Stone, Sophia Shao,
Anitha Vijayakumar, Shibo Wang,                            
Qiumin Xu,
Doe Hyun Yoon, Cliff Young
for their support and feedback.

\bibliographystyle{IEEEtranS}


\end{document}